\documentclass[10pt,twocolumn,letterpaper]{article}
\usepackage{cvpr}
\usepackage[dvipsnames]{xcolor}

\usepackage{graphicx}
\usepackage{amsmath}
\usepackage{amssymb}
\usepackage{booktabs}
\usepackage{enumitem}
\usepackage{microtype}
\usepackage[dvipsnames]{xcolor}
\usepackage{bm, bbm}
\usepackage{algorithmic}
\usepackage[linesnumbered,boxed,ruled,commentsnumbered]{algorithm2e}
\usepackage{multirow}
\usepackage{colortbl}
\definecolor{skyblue}{RGB}{0,200,255}

\newcommand{\hlr}[1]{{\color{red}{#1}}}

\newcommand{\hly}[1]{{\color{yellow}{#1}}}

\usepackage{bbm}
\usepackage{amsmath}
\usepackage[normalem]{ulem}
\useunder{\uline}{\ul}{}

\usepackage{lipsum}
\newcommand\blfootnote[1]{%
\begingroup
\renewcommand\thefootnote{}\footnote{#1}%
\addtocounter{footnote}{-1}%
\endgroup
}

\definecolor{cvprblue}{rgb}{0.21,0.49,0.74}
\usepackage[pagebackref,breaklinks,colorlinks,citecolor=cvprblue]{hyperref}

\title{
HiP-AD: Hierarchical and Multi-Granularity Planning with Deformable Attention for Autonomous Driving in a Single Decoder
}

\author{Yingqi Tang$^{\star}$ \quad Zhuoran Xu$^{\star}$ \quad Zhaotie Meng \quad Erkang Cheng$^{\boxtimes}$\\	
[2mm]
~\normalsize{Nullmax} \quad	\\	
{\tt\small \{tangyingqi, xuzhuoran, mengzhaotie, chengerkang\}@nullmax.ai}\\
\normalsize{
\url{https://github.com/nullmax-vision/HiP-AD}
}
}

\begin{document}
\maketitle
\blfootnote{$^\star$ Equal contribution; $^\boxtimes$ Corresponding author.}
\begin{abstract}
Although end-to-end autonomous driving (E2E-AD) technologies have made significant progress in recent years, there remains an unsatisfactory performance on closed-loop evaluation. 
The potential of leveraging planning in query design and interaction has not yet been fully explored.
In this paper, we introduce a multi-granularity planning query representation that integrates heterogeneous waypoints, including spatial, temporal, and driving-style waypoints across various sampling patterns. It provides additional supervision for trajectory prediction, enhancing precise closed-loop control for the ego vehicle.
Additionally, we explicitly utilize the geometric properties of planning trajectories to effectively retrieve relevant image features based on physical locations using deformable attention. 
By combining these strategies, we propose a novel end-to-end autonomous driving framework, termed HiP-AD, which simultaneously performs perception, prediction, and planning within a unified decoder.
HiP-AD enables comprehensive interaction by allowing planning queries to iteratively interact with perception queries in the BEV space while dynamically extracting image features from perspective views.
Experiments demonstrate that HiP-AD outperforms all existing end-to-end autonomous driving methods on the closed-loop benchmark Bench2Drive and achieves competitive performance on the real-world dataset nuScenes. 

\vspace{-0.5cm}

\end{abstract}

\section{Introduction}
\label{sec:intriduction}

\begin{figure}[htbp]
  \centering
  \includegraphics[width=0.88\linewidth]{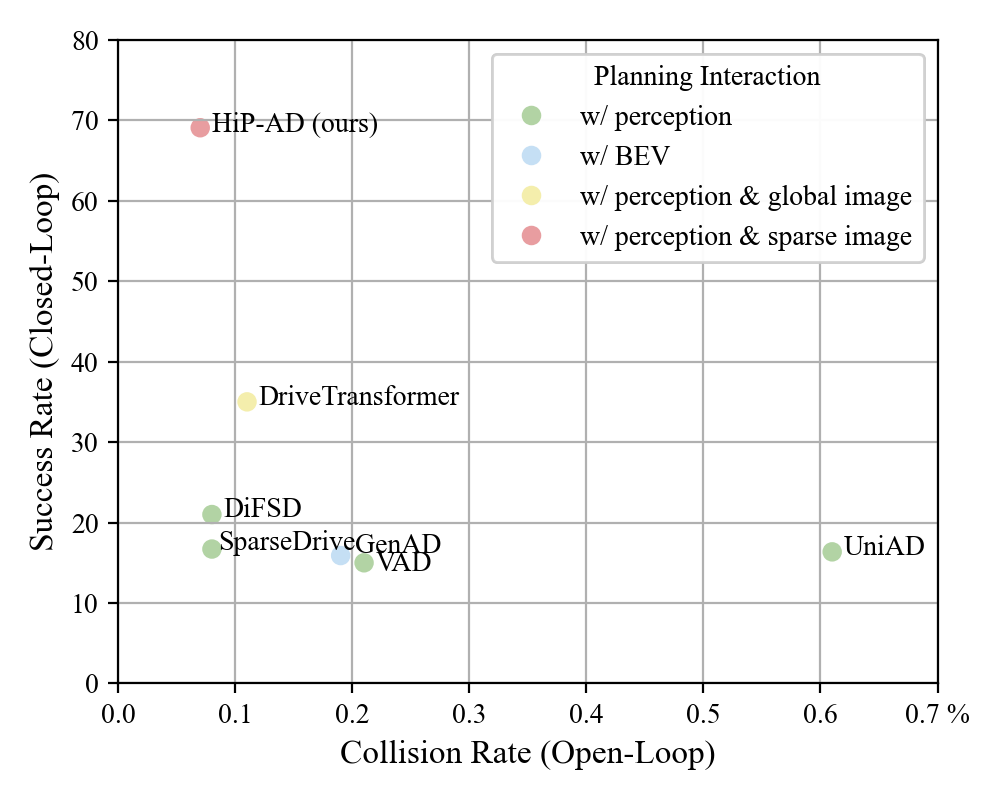}
  \caption{Comparison of existing state-of-art works on open-loop metric of Collision Rate on nuScenes dataset and closed-loop metric of Success Rate on Bench2Drive dataset, where \textbf{top left is better}. The legend indicates different planning interaction methods.}
  \label{fig:1_score_scatter}
\vspace{-0.5cm}
\end{figure}

\begin{figure}[t]
  \centering
  \includegraphics[width=0.85\linewidth]{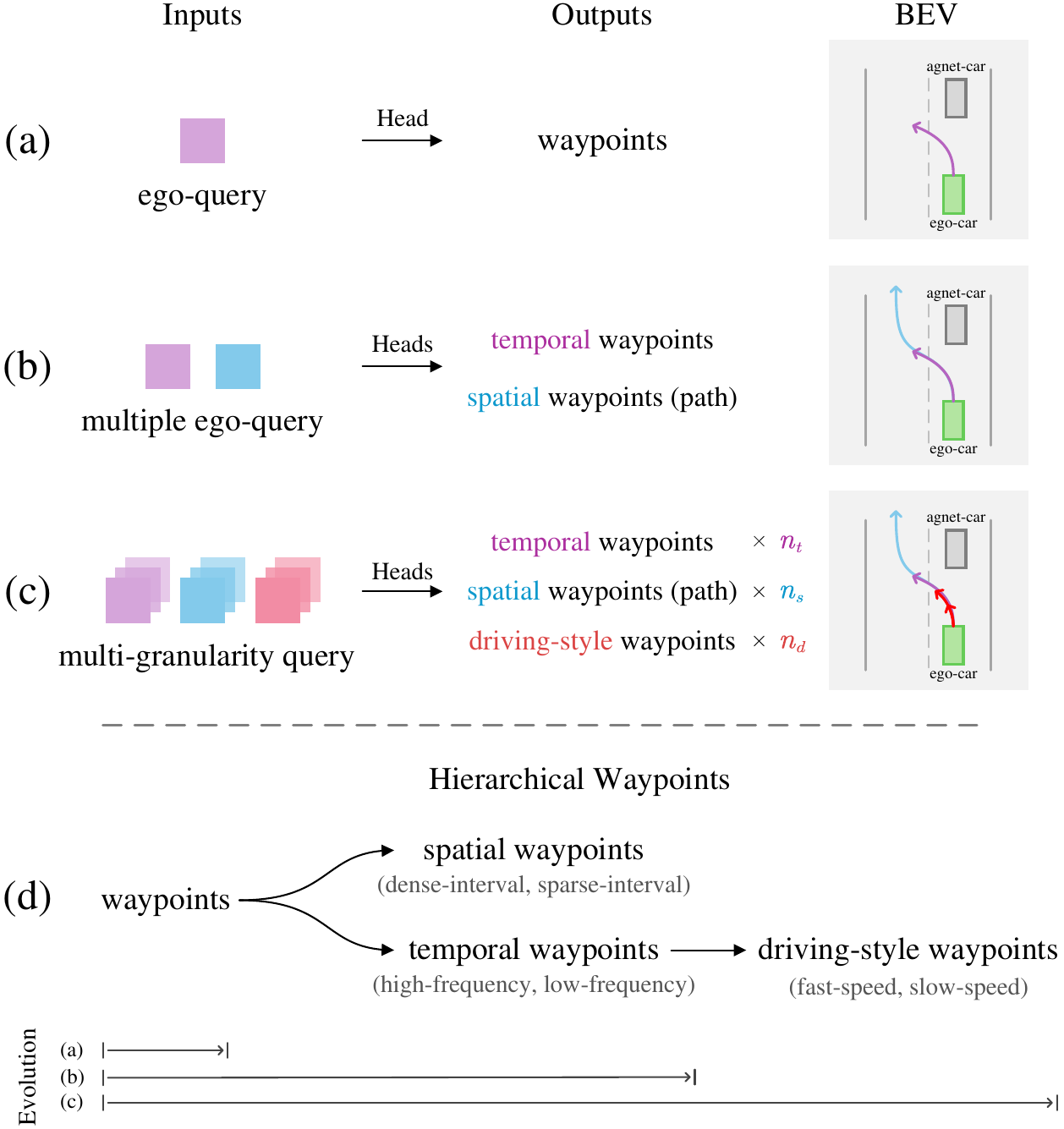}
  \caption{This diagram compares earlier methods (a-b) for predicting waypoints with our proposed multi-granularity planning design (c), where $n_t$, $n_s$, and $n_d$ represent different number of granularity in each waypoints type in terms of frequency, interval, and speed. Part (d) illustrates the evolution of hierarchical waypoints with instantiated granularity based on different sampling strategies.}
  \label{fig:2_comparasion}
  \vspace{-0.5cm}
\end{figure}

Recently, great progress has been achieved in end-to-end autonomous driving (E2E-AD), which directly predicts planning trajectory from raw sensor data.
One of the mainstream methods is to integrate all tasks (e.g., perception, prediction, and planning) into a single model within a fully differentiable manner~\cite{UniAD, VAD, PARA-Drive}.
Compared to the traditional standalone or multi-task paradigm~\cite{treiber2000congested, dauner2023parting, pini2023safe}, it greatly alleviates the accumulative errors and makes all task modules work collaboratively, which exhibits promising performance under the effect of large-scale data.

Despite these advances, a significant performance gap persists between open-loop and closed-loop evaluations, primarily attributable to differences in motivations.
Open-loop methods focus on the displacement error in the planning trajectory compared to the ground truth, while closed-loop methods prioritize safe driving performance. 
As illustrated in \cref{fig:1_score_scatter}, previous E2E-AD methods~\cite{UniAD, VAD, SparseDrive, GenAD, DiFSD} demonstrate strong performance in terms of Collision Rate (lower is better) in the open-loop benchmark nuScenes~\cite{nuscenes} with some methods achieving as low as 0.1\%. 
However, these methods show unsatisfactory performance in terms of Success Rate on the comprehensive closed-loop evaluation dataset, Bench2Drive~\cite{bench2drive}, where the Success Rate remains below 35\%.
Even when focusing solely on emergency braking, the Success Rate is still inadequate, below 55\% (as shown in Tab.~\ref{tab:bench2dirve2}), despite achieving an open-loop Collision Rate as low as 0.1\%. 

We argue that the potential of planning in query design and interaction has not been fully explored in these E2E-AD methods.
First, most methods~\cite{UniAD, VAD, SparseAD, PARA-Drive, DriveTransformer} formulate E2E-AD as an imitation learning task via a trajectory regression (~\cref{fig:2_comparasion} (a)) with sparse supervision, focusing primarily on the trajectory fitting itself rather than closed-loop control.
In contrast, closed-loop oriented methods~\cite{codevilla2019exploring, prakash2021multi, shao2022interfuser, Jaeger2023HiddenBO} encounter several other challenges, such as non-convex problems~\cite{VADv2} and steering errors~\cite{CarLLaVA}.
CarLLaVA~\cite{CarLLaVA} greatly alleviates these issues by decoupling standard waypoints into time-conditioned and space-conditioned waypoints for longitudinal and lateral control, as shown in~\cref{fig:2_comparasion} (b).
However, it is built upon a pre-trained large language model without intermediate perceptual results, which lacks interpretability, and it has not investigated the diversification of trajectories.
In this paper, we propose multi-granularity planning query representation with hierarchical waypoints predictions for E2E-AD, as shown in~\cref{fig:2_comparasion} (c-d). 
Specifically, we disentangle the waypoints into temporal, spatial (path), and driving-style waypoints predictions with corresponding planning queries\footnote{some works refer to this as ego query}. 
Additionally, we further diversify each type of waypoints into multiple granularities with different sampling strategies, such as frequency, distance, and speed, enriching additional supervision during training.
They can be effectively aggregated to facilitate interaction between the different characteristics.
As a result, sparse waypoints provide global information, and dense waypoints outputs are more suitable for fine-grained control.
Moreover, the multi-granularity dramatically reduces the ego hesitation issue that the ego vehicle keeps waiting in some scenarios until the closed-loop simulation runs out of time. 
It encourages behavior learning in complex scenarios (e.g., traffic signs, overtaking) without introducing causal clues.

Second, sequential paradigms like UniAD~\cite{UniAD} and VAD~\cite{VAD} formulate interaction only between learnable ego queries and the perception transformer outputs.
While the parallel approach of Para-Drive~\cite{PARA-Drive} applies the ego query interacting solely with BEV features. These approaches lack the comprehensive interaction for planning to effectively engage with both perception and scene features (e.g., image or BEV features).
In contrast, DriveTransformer~\cite{DriveTransformer} allows ego query to fully interact with both perception and image features within a single Transformer.
However, it remains challenging for the ego query to effectively extract valuable information from multi-view images throgh global attention without the prior context of the planning trajectory.
To address this issue, we employ planning deformable attention with physical locations to dynamically sample relevant image features in proximity to the planning trajectory.
It can be easily integrated into a unified framework alongside perception tasks.
Specifically, we employ a unified decoder with hybrid task queries as inputs, which combines queries from detection tasks (including motion prediction), map understanding tasks, and planning tasks. 
It enables planning and perception tasks to exchange information in the BEV space and allows planning query to interact with the image space by leveraging the geometric priors of their waypoints. 

With both strategies, we propose a novel end-to-end autonomous driving framework, termed HiP-AD. 
It is evaluated the on closed-loop benchmark Bench2Drive \cite{bench2drive} and open-loop dataset nuScenes \cite{nuscenes}, achieving outstanding results on both planning and perception tasks.
Our main contributions are as follows:

\begin{itemize}
    \item We propose a multi-granularity planning query representation that integrates various characteristics of waypoints to enhance the diversity and receptive field of the planning trajectory, enabling additional supervision and precise control in a closed-loop system.
    \item We propose a planning deformable attention mechanism that explicitly utilizes the geometric context of the planning trajectory, enabling dynamic retrieving image features from the physical neighborhoods of waypoints to learn sparse scene representations.
    \item We propose a unified decoder where a comprehensive interaction is iteratively conducted on planning-perception and planning-images, making an effective exploration of end-to-end driving in both BEV and perspective view, respectively. 
\end{itemize}

\begin{figure*}[ht!]
  \centering
  \includegraphics[width=0.95\textwidth]{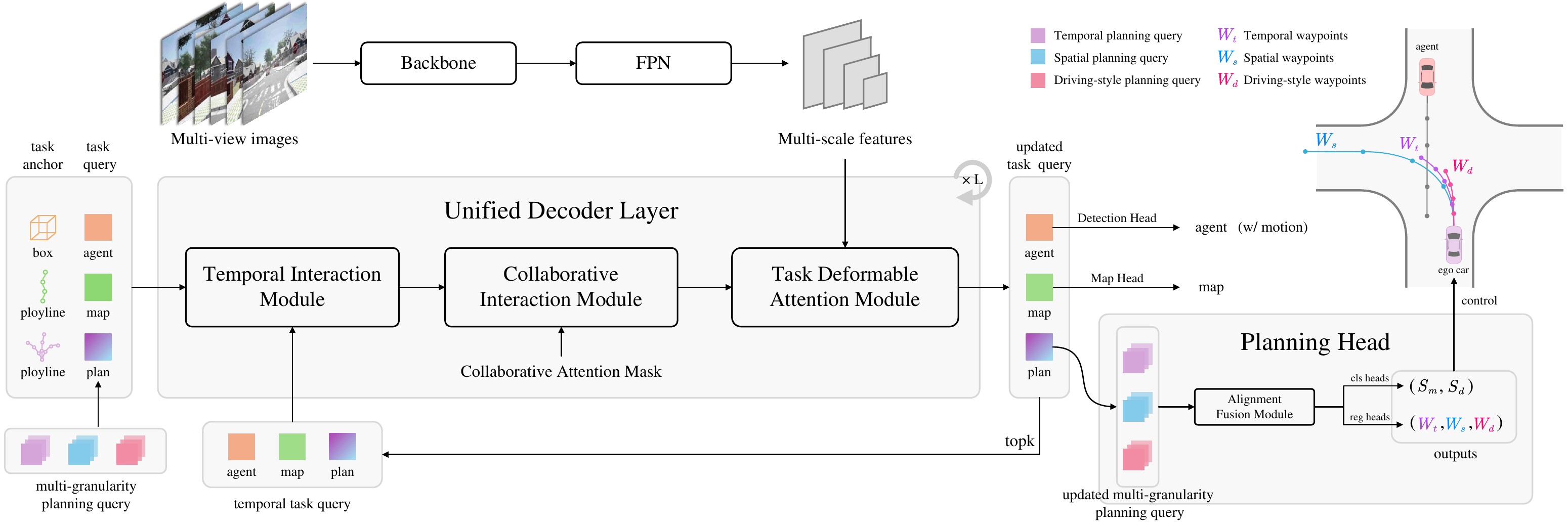}
  \caption{\textbf{The overall framework of HiP-AD}. It consists of a Backbone and a FPN for extracting image features, a unified decoder for iteratively updating query, and various heads for multi-task prediction. 
  The inputs of the unified decoder are task anchors and queries (agent, map, and planning), where planning query consist of multi-granularity waypoints representations.
  In each unified decoder layer, the task queries first interact with temporal query separately, then collaboratively with each other, and finally engage the image features in an iterative manner.
  Last, the updated task queries are sent to the corresponding heads for perception, prediction, and planning, where planning results including various waypoints with different granularity for precise action control.
  }
  \label{fig:3_framework}
  \vspace{-0.3cm}
\end{figure*}

\section{Related Work}

\subsection{Dynamic and Static Perception}
Dynamic object detection can be broadly categorized into BEV-based methods~\cite{LSS, BEVFormer, BEVDet, BEVDepth} and sparse query-based methods~\cite{PETR, StreamPETR, DETR3D, Sparse4D}.
The BEV-based methods detect objects by constructing bird's-eye view (BEV) features, either by flattening voxel features with estimated depth or by injecting perspective features into BEV grids. In contrast, sparse query-based methods directly compute detection results through techniques such as geometric embedding and anchor projection with deformable attention, reducing the need for extensive construction of BEV features and leading to a more efficient detection process.

For static map elements detection tasks, the dominant BEV representation includes two distinct map approaches: rasterized-based methods and vectorized-based methods.
Rasterized-based methods~\cite{HDMapNet} predict static elements through techniques such as segmentation and lane detection, followed by a post-processing step. In contrast, vectorized representation methods~\cite{VectorMapNet, MapTR} directly model vectorized instances without post-processing, leading to improved efficiency and higher performance. Building on this paradigm, advancements such as long-sequence temporal fusion~\cite{StreamMapNet} and multi-points scatter strategies~\cite{MapQR} further enhance the robustness of static element detection.

\vspace{-0.05cm}
\subsection{Motion Prediction and Planning}
Motion prediction methods predict vehicle trajectories from LiDAR points or rendered HD maps.
For example, IntentNet~\cite{Intentnet} develops a one-stage detector and forecaster with 3D point clouds and dynamic maps.
VectorNet~\cite{VectorNet} proposes an efficient graph neural network for predicting the behaviors of multiple agents.
PnPNet~\cite{PnPNet} develops a multi-object tracker that simultaneously performs perception and motion forecasting, leveraging rich temporal context for enhanced performance.
ViP3D~\cite{ViP3D} proposes a fully differentiable vision-based trajectory prediction approach.

Traditional planning has often been treated as an independent task, relying on the outputs from perception and prediction modules.
Early works utilize rule-based planners~\cite{treiber2000congested,bouchard2022rule,dauner2023parting} or optimization-based methods~\cite{fan2018baidu,ziegler2014trajectory,shi2022path} for planning.
In contrast, learning-based methods have gained prominence, offering new paradigms for planning tasks.
Some works~\cite{codevilla2019exploring,pomerleau1988alvinn,prakash2021multi,codevilla2018end} directly predict planning trajectories or control signals without intermediate results, while other methods~\cite{cui2021lookout,sadat2020perceive} optimize planning by utilizing outputs of perception and prediction for great interpretability.
Additionally, reinforcement learning~\cite{kendall2019learning,chen2021learning,chekroun2023gri,toromanoff2020end} has shown significant promise in autonomous driving.

\vspace{-0.1cm}
\subsection{End-to-end Autonomous Driving}
Different from the traditional standalone or multi-task paradigm \cite{treiber2000congested, dauner2023parting, pini2023safe}, End-to-End autonomous driving methods alleviate potential cumulative errors by applying a unified pipeline that integrates perception, prediction, and planning tasks.
For example, UniAD~\cite{UniAD} pioneeringly integrates various tasks into a single model.
VAD~\cite{VAD,VADv2} simplifies the scene representation to vectorized elements, enhancing both efficiency and robustness.
PPAD~\cite{PPAD} formulates a hierarchical dynamic key object attention to model the interactions in an interleaving and autoregressive manner.
Instead of constructing dense BEV features, SparseAD~\cite{SparseAD} and SparseDrive~\cite{SparseDrive} utilize a sparse query-based framework, achieving greater efficiency and more accurate results.
Based on this manner, DiFSD~\cite{DiFSD} iteratively refines the ego trajectory from coarse-to-fine through geometric information.
In contrast to the sequential scheme, Para-Drive~\cite{PARA-Drive} employs a parallel approach, performing perception and planning tasks concurrently for improved performance.
DriveTransformer~\cite{DriveTransformer} further integrates these tasks within a single transformer and achieves excellent performance in closed-loop system.
Additionally, generative frameworks~\cite{GenAD, DiffusionDrive} and large language models~\cite{CarLLaVA, LMDrive, VLP} provide innovative perspectives for decision-making and planning, and are increasingly capturing attention in the field.
\section{Method}
\subsection{Overview}
\label{overview}

The overall network architecture of HiP-AD is illustrated in \cref{fig:3_framework}.
It consists of a backbone followed by a feature pyramid network (FPN) module that extracts multi-scale features $\{F_i\}_{i=1}^V$ from multi-view images $\{I_i\}_{i=1}^V $, and a unified decoder with various task-specific heads.
The unified decoder takes hybrid task anchors and queries as inputs, which are concatenated from agent queries $Q_{a} \in \mathbb{R}^{N_{a}\times C}$, map queries $Q_{m} \in \mathbb{R}^{N_{m}\times C}$, and planning queries $Q_{p} \in \mathbb{R}^{N_{p}\times C}$, where $N$ represents the number of queries and $C$ denotes the feature channel size.
The agent query corresponds to object detection and motion prediction, while the map and multi-granularity planning queries (\cref{multi-granularity_planning}) manage online mapping and trajectory prediction.
The detection and motion prediction heads, along with the map and planning heads, predict their respective tasks. 
The planning head outputs temporal, spatial, and driving-style waypoints for ego-vehicle control. Additionally, the top $k_{a}, k_{m}, k_{p}$ updated queries are stored in memory for subsequent temporal interactions.

\begin{figure}[t]
  \centering
  \includegraphics[width=0.94\linewidth]{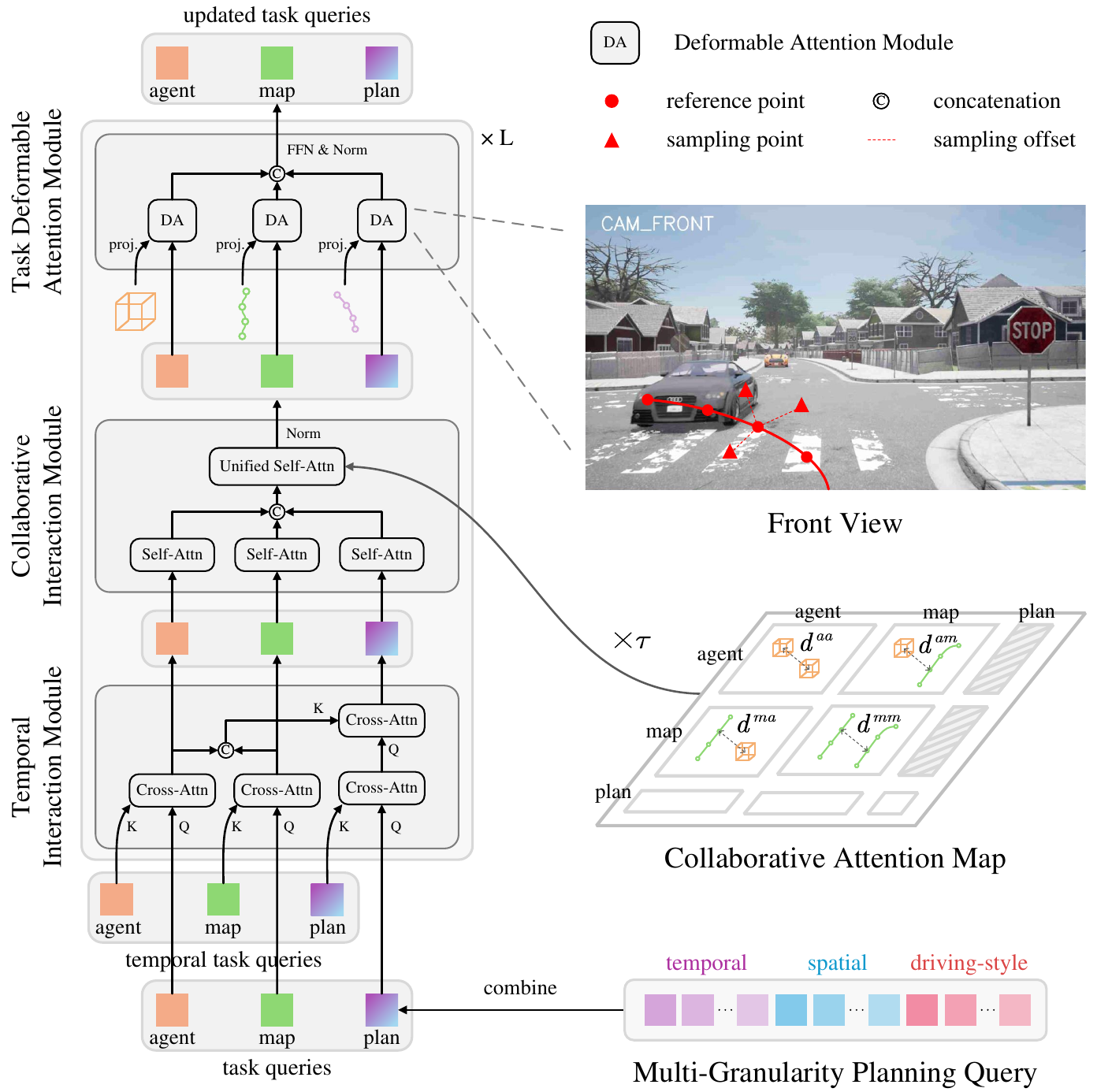}
  \vspace{-0.1cm}
  \caption{Illustration of the detailed architecture of three submodules within the unified decoder layer for comprehensive interaction.}
  \label{fig:4_unified_decoder}
  \vspace{-0.5cm}
\end{figure}

\subsection{Unified Decoder}

As shown in \cref{fig:3_framework} and in detail in \cref{fig:4_unified_decoder}, the unified decoder consists of three modules: the Temporal Interaction Module, the Collaborative Interaction Module, and the Task Deformable Aggregation Module. Each module is designed to facilitate temporal, cross-task, and task-image interactions, respectively.
Each input task query is associated with a corresponding anchor. The agent query uses box anchors, $A_{a} \in \mathbb{R}^{N_{a} \times D_{a}}$, while the map query utilizes polyline anchors, $A_{m} \in \mathbb{R}^{N_{m} \times D_{m}}$, initialized through a clustering algorithm, where $D$ denotes the anchor dimension.
We also model the planning query as a polyline anchor $A_{p}\in\mathbb{R}^{N_{p} \times D_{p}}$ using its $T$ future waypoitns. 

\noindent
\textbf{Temporal Interaction.}
The Temporal Interaction Module establishes communication between the features of the current task and those of historical tasks, which are retained from the previous inference frame via a top $k$ selection mechanism.
As shown in the bottom left of the~\cref{fig:4_unified_decoder}, there are three distinct cross-attention mechanisms for each task's temporal interaction, as well as an additional cross-attention mechanism that enables enhanced interaction between the planning query and temporal perception queries, focusing on historical surrounding elements.

\noindent
\textbf{Collaborative Interaction.}
The Collaborative Interaction Module enables cross-task interaction. It incorporates three separate self-attention mechanisms, each dedicated to an individual task, as well as a unified self-attention module to apply interaction across tasks.
Instead of using global attention, we construct a geometric attention map for each query pair to focus on local and relative elements. 
Using perception query as an example, following~\cite{SparseBEV}, we dynamically adjust BEV receptive fields through scaling distance as attention weight.
\begin{equation}
\text{Attention}(Q, K, V) = \text{Softmax}(\frac{QK^T}{\sqrt{C}}-\tau D)V,
\end{equation}
where $\tau$ is a learnable coefficient computed from $Q$ by MLP functions. $D$ represents the Euclidean distance between two object instances, denoted as ($d_{i,j}$).
Similarly, we extend the computation of minimum distances to generate the attention weight by incorporating interactions between map-agent, map-map, and agent-map anchors.
For planning queries, there are no distance constraints, allowing them to access information from all tasks. 

\noindent
\textbf{Task Deformable Attention.}
Unlike previous works~\cite{DriveTransformer}, which employ global attention to interact with all multi-view image features, we leverage separate deformable attention modules to sample local sparse features tailored to each task query.
Specifically, we project task anchors to multi-view images through the camera parameters, as used in~\cite{Sparse4Dv2, SparseDrive}.
For planning, we distribute reference waypoints across various predefined height values and then project them onto multi-view images.
To sample features of neighboring points, we employ several MLPs to learn spatial offsets and associated weights based on the projected reference points.
The process of planning deformable attention (PDA) can be formulated as:
\begin{equation}
\text{PDA}(Q_{p}, F) = \sum_{i\in V} \text{DeformAttn}(Q_{p}, \mathcal{P}(A_{p}), F_i),
\end{equation}
where $\mathcal{P}$ indicates a project function.
Therefore, it integrates features around the future trajectory to learn the sparse scene representation, avoiding a potential collision.

\subsection{Hierarchical and Multi-granularity Planning}
\label{multi-granularity_planning}

\noindent
\textbf{Hierarchical Waypoints.}
Different from previous waypoints designs~\cite{VAD, CarLLaVA}, we not only utilize temporal and spatial waypoints, but also introduce driving-style waypoints, as shown in~\cref{fig:2_comparasion} (d). While similar to temporal waypoints, driving-style waypoints further integrate the velocity to learn ego action in complex environment.
Additionally, a multi-sampling strategy is applied to enable rich trajectory supervision and precise control.
This strategy combines dense and sparse intervals for spatial waypoints, high and low frequencies for temporal and driving-style waypoints, and further incorporates varying speeds for driving-style waypoints.

As a result, sparse-interval waypoints provide a broader global context, which aids in advanced decision-making while dense-interval waypoints enable fine-grained control for precise maneuvering.
The temporal and spatial waypoints, along with high-low frequency and dense-sparse interval sampling strategies, complementarily contribute to more robust and effective planning.
Moreover, driving-style waypoints with different speeds provide a rich understanding of scenarios such as overtaking or emergency braking, thereby providing flexible longitudinal control in close-loop evaluation.

\begin{figure}[t]
  \centering
  \includegraphics[width=0.94\linewidth]{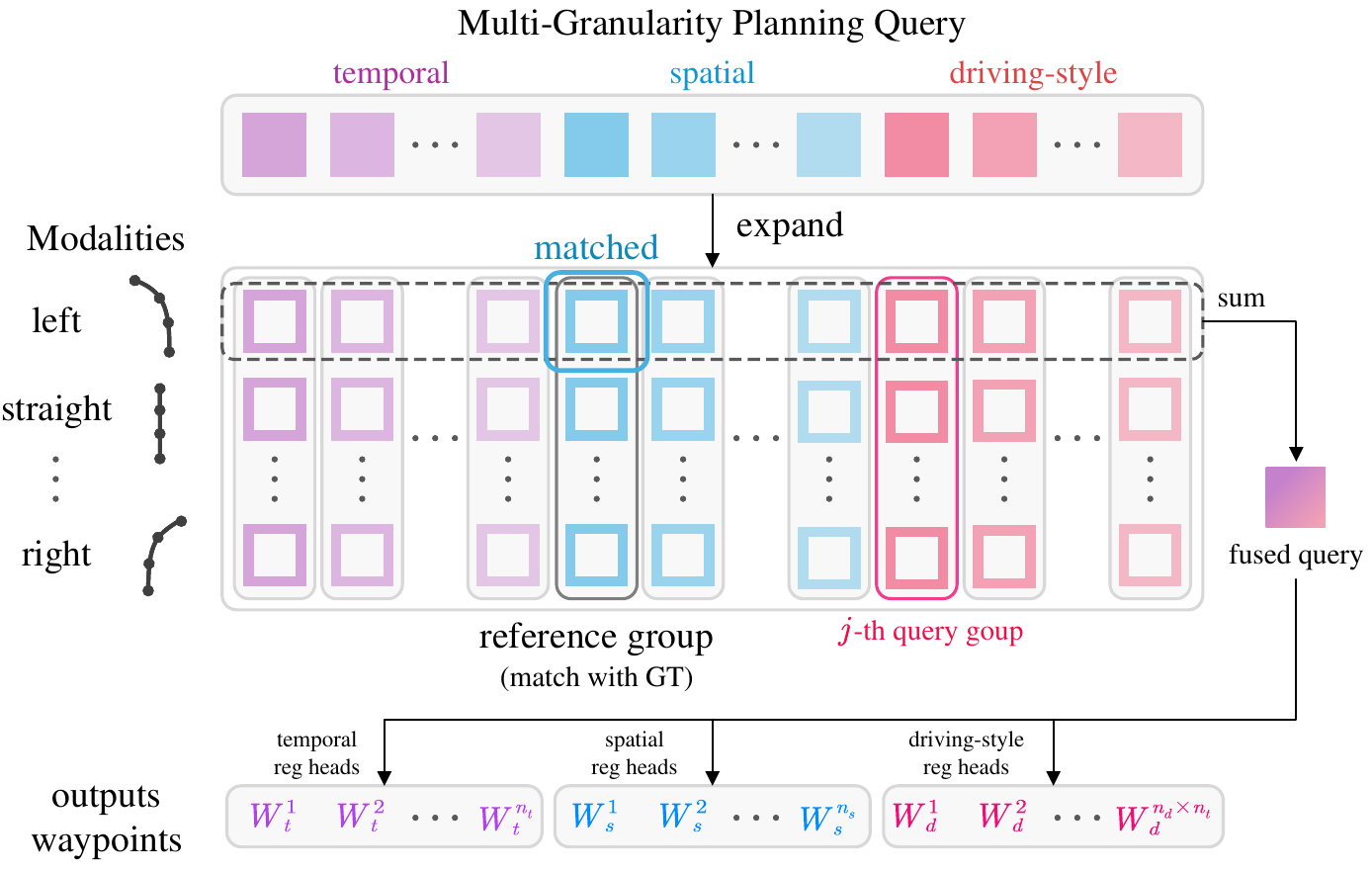}
  \vspace{-0.2cm}
  \caption{The illustration of multi-granularity query architecture with alignment fusion for waypoints prediction. We omit the classification head for clarity.}
  \label{fig:5_granularity_fusion}
  \vspace{-0.5cm}
\end{figure}

\noindent
\textbf{Multi-granularity Planning Query.}
We construct multi-granularity planning queries to predict these heterogeneous waypoints. 
As illustrated in~\cref{fig:5_granularity_fusion}, there are $N_g$ granularity, which consist of temporal, spatial, and driving-style planning queries with $n_t$, $n_s$, $n_d \times n_t$ sampling strategies.
Each granularity planning query includes $N_m$ modality, representing trajectories like left, straight, right, etc.
The total number of multi-granularity planning queries is given by  $N_{p} = N_m \times N_g $, where $N_g = n_t + n_s + n_d \times n_t$.

Upon processing through the unified decoder, planning queries of varying granularities within a single modality are aligned and aggregated to create a fused query, enhancing information complementarity and overall effectiveness.
The fused query is employed to predict waypoints across all granularities, leveraging additional supervision to optimize trajectory.
The process is formulated as follows:
\begin{equation}
Q_{fuse}^{i} = \sum_{j \in N_{g}}Q_{p}^{ij},
\end{equation}
where $i$, $j$ are the $i$-th modality and $j$-th granularity, respectively.
$Q_{fuse}^{i}$ represents the fused query of $i$-th modality.
We use $N_{g}$ $\text{MLP}_{reg}$ layers to regress different granularity of waypoints $W^{i,j}$, while all granularity share the same modality score layer $\text{MLP}_{cls\text{-}m}$,
\begin{equation}
\begin{split}
W^{i,j} = \text{MLP}^{j}_{reg}(Q_{fuse}^{i})& \\
S^i_{m} = \text{MLP}^{i}_{cls\text{-}m}(Q_{fuse}^{i})&,
\end{split}
\end{equation}
where modality score $S_m$ is used in the inference step to select the best modality.
Additionally, a driving-style classification head is used for final waypoints selection.
\begin{equation}
S^{i}_{d} = \text{MLP}^{i}_{cls\text{-}d}(Q_{fuse}^{i}).
\end{equation}

\begin{table*}[tb]\footnotesize
\centering
\scalebox{0.83}
{
\begin{tabular}{l|c|cccc}
\toprule
\multirow{2}{*}{\textbf{Method}}  & \textbf{Open-loop Metric} & \multicolumn{4}{c}{\textbf{Closed-loop Metric}}  \\ \cmidrule{2-6} 
                                 &  Avg. L2 $\downarrow$ & Driving Score $\uparrow$  & Success Rate(\%) $\uparrow$ & Efficiency $\uparrow$ & Comfortness $\uparrow$\\ \midrule
AD-MLP~\cite{ADMLP}              & 3.64 & 18.05 & 0.00  & 48.45  & 22.63   \\ 
UniAD-Tiny~\cite{UniAD}          & 0.80 & 40.73 & 13.18 & 123.92 & 47.04   \\
UniAD-Base~\cite{UniAD}          & 0.73 & 45.81 & 16.36 & 129.21 & 43.58   \\
VAD~\cite{VAD}                   & 0.91 & 42.35 & 15.00 & 157.94 & 46.01   \\
SparseDrive~\cite{SparseDrive}   & 0.87 & 44.54 & 16.71 & 170.21 & \textbf{48.63}   \\ 
GenAD~\cite{GenAD}               & -    & 44.81 & 15.90 & -      & -       \\ 
DiFSD~\cite{DiFSD}               & 0.70 & 52.02 & 21.00 & 178.30 & -       \\
DriveTransformer-Large~\cite{DriveTransformer} 
                                 & \textbf{0.62} & 63.46 & 35.01 & 100.64 & 20.78   \\
\midrule
TCP-traj*~\cite{wu2022trajectoryguided}            
                                 & 1.70 & 59.90 & 30.00 & 76.54  & 18.08   \\
ThinkTwice*~\cite{jia2023thinktwice}          
                                 & 0.95 & 62.44 & 31.23 & 69.33  & 16.22   \\
DriveAdapter*~\cite{jia2023driveadapter}        
                                 & 1.01 & 64.22 & 33.08 & 70.22  & 16.01   \\
\midrule                                
HiP-AD                           & 0.69 & \textbf{86.77} & \textbf{69.09} & \textbf{203.12} & 19.36   \\
\bottomrule
\end{tabular}
}
\caption{Open-loop and Closed-loop results of planning in Bench2Drive. Avg. L2 is averaged over the predictions in 2 seconds under 2Hz. * denotes expert feature distillation. }
\label{tab:bench2dirve1}
\vspace{-0.2cm}
\end{table*}

\begin{table*}[tb]\footnotesize
\centering
\scalebox{0.85}
{
\begin{tabular}{l|ccccc|c}
\toprule
\multirow{2}{*}{\textbf{Method}} & \multicolumn{5}{c}{\textbf{Ability} (\%) $\uparrow$}  \\ \cmidrule{2-7} 
& \multicolumn{1}{c}{Merging} & \multicolumn{1}{c}{Overtaking} & \multicolumn{1}{c}{Emergency Brake} & \multicolumn{1}{c}{Give Way} & Traffic Sign & \textbf{Mean} \\ \midrule
AD-MLP~\cite{ADMLP}            & 0.00  & 0.00  & 0.00  & 0.00  &  4.35 & 0.87   \\
UniAD-Tiny~\cite{UniAD}        & 8.89  & 9.33  & 20.00 & 20.00 & 15.43 & 14.73  \\ 
UniAD-Base~\cite{UniAD}        & 14.10 & 17.78 & 21.67 & 10.00 & 14.21 & 15.55  \\ 
VAD~\cite{VAD}                 & 8.11  & 24.44 & 18.64 & 20.00 & 19.15 & 18.07  \\ 
DriveTransformer-Large~\cite{DriveTransformer}  
                               & 17.57 & 35.00 & 48.36 & 40.00 & 52.10 & 38.60  \\ 
\midrule
TCP-traj*~\cite{wu2022trajectoryguided}            
                               & 12.50 & 22.73 & 52.72 & 40.00 & 46.63 & 34.92  \\
ThinkTwice*~\cite{jia2023thinktwice}          
                               & 13.72 & 22.93 & 52.99 & \textbf{50.00} & 47.78 & 37.48  \\
DriveAdapter*~\cite{jia2023driveadapter}        
                               & 14.55 & 22.61 & 54.04 & \textbf{50.00} & 50.45 & 38.33  \\

\midrule                                
HiP-AD                         & \textbf{50.00} & \textbf{84.44} & \textbf{83.33} & 40.00 & \textbf{72.10} & \textbf{65.98}  \\ 

\bottomrule
\end{tabular}
}
\caption{Multi-Ability Results of E2E-AD Methods in Bench2Drive.* denotes expert feature distillation.}
\label{tab:bench2dirve2}
\vspace{-0.3cm}
\end{table*}

\noindent
\textbf{Align Matching.}
\label{sampling_matching}
In the training process, a winner-takes-all matching approach is employed within each query group, which includes all modalities with a specific granularity, to select the optimal modality for optimization.
Instead of performing independent matches for each group of waypoints, we introduce an align matching strategy that designates a single group of waypoints as the reference waypoints $W^{i,\text{ref}}$ along with its corresponding GroundTruth $GT^{\text{ref}}$ for matching.
\begin{equation}
\label{eq:align-matching}
\mathbbm{1}_{\text{ref}} = \mathop{\arg\min}_i(L_2(W^{i,\text{ref}}-GT^{\text{ref}})),
\end{equation}
where ${\mathbbm{1}_{\text{ref}}}$ represents the best matching index in reference query group.
All other groups then share the same matching results to align with the matched planning modalities on rest query groups.
Consequently, the gradients across all granularities of the optimally matched modality can be effectively backpropagated.

Based on this mechanism, $n_d$ driving-style waypoints are selected after the align-matching.
Each driving-style waypoints is responsible for an area of velocity.
Different from spatial or temporal waypoints that optimize across all granularities, we only select one granularity of driving-style waypoints to optimize, ensuring that each granularity of waypoints learns corresponding actions to various complex driving scenarios.
This process is informed by the ground truth of the temporal waypoints.

\subsection{Waypoints Selection and Action Control}
\noindent
\textbf{Selection.}
During the inference step, the final waypoints are computed through a two-step selection process. First, the optimal modality is selected based on the predicted modality score $S_m$. 
Second, specific granularity waypoints are chosen according to predefined rules: dense intervals are selected for spatial waypoints, and high-frequency waypoints are preferred for temporal granularity. For driving-style waypoints, the selection is based on the highest score from the predicted style classifications.

\begin{table*}[t]\footnotesize
\centering
\scalebox{0.87}
{
\begin{tabular}{l|cccc|cccc|c|c}
\toprule[1pt]
\multirow{2}{*}{\textbf{Method}} & 
\multicolumn{4}{c|}{\textbf{L2 (\textit{m}) $\downarrow$}} &
\multicolumn{4}{c|}{\textbf{Collision (\%)} $\downarrow$}  &
\multirow{2}{*}{\textbf{Latency} $\downarrow$}  &
\multirow{2}{*}{\textbf{FPS} $\uparrow$} \\

& 
\multicolumn{1}{c}{1\textit{s}} &
\multicolumn{1}{c}{2\textit{s}} &
\multicolumn{1}{c}{3\textit{s}} &
\cellcolor[HTML]{DADADA}Avg.    &
\multicolumn{1}{c}{1\textit{s}} &
\multicolumn{1}{c}{2\textit{s}} &
\multicolumn{1}{c}{3\textit{s}} &
\cellcolor[HTML]{DADADA}Avg.    &
&  
\\ 
\midrule

VAD-Base~\cite{VAD} &  0.41 & 0.70 & 1.05 & \cellcolor[HTML]{DADADA}0.72 & 0.03 & 0.19 & 0.43 & \cellcolor[HTML]{DADADA}0.21 & 224.3 & 4.5 \\
GenAD~\cite{GenAD}    &  0.28 & 0.49 & 0.78 & \cellcolor[HTML]{DADADA}0.52 & 0.08 & 0.14 & 0.34 & \cellcolor[HTML]{DADADA}0.19 & 149.2 & 6.7 \\ 
SparseDrive-S~\cite{SparseDrive}  &  0.29 & 0.58 & 0.96 & \cellcolor[HTML]{DADADA}0.61 & 0.01 & 0.05 & 0.18 & \cellcolor[HTML]{DADADA}0.08 & 111.1 & 9.0 \\ 
DiFSD-S~\cite{DiFSD} & 0.15 & 0.31 & 0.56 & \cellcolor[HTML]{DADADA}0.33  & 0.00 &  0.06 & 0.19 & \cellcolor[HTML]{DADADA}0.08 & 93.7 & 10.7 \\ 
DriveTransformer-Large~\cite{DriveTransformer}& 0.16 & 0.30 & 0.55 & \cellcolor[HTML]{DADADA}0.33  & 0.01 & 0.06 & 0.15 & \cellcolor[HTML]{DADADA}0.07 & 221.7 & 4.5 \\ 
\midrule 
HiP-AD  & 0.28 & 0.53 & 0.87 & \cellcolor[HTML]{DADADA}0.56 & 0.01 & 0.05 & 0.15 & \cellcolor[HTML]{DADADA}0.07 & 109.9 & 9.1 \\

\bottomrule
\end{tabular}
}
\caption{Open-loop planning evaluation results on the nuScenes validation dataset with the evaluation protocol~\cite{li2024ego}. Latency and FPS of DriveTransformer~\cite{DriveTransformer} are measured on NVIDIA H800 GPU while the others are measured on NVIDIA 3090 GPU.}
\label{tab:nuscenes1}
\vspace{-0.2cm}
\end{table*}
\begin{table}[t]\footnotesize
\centering
\scalebox{0.87}
{
\begin{tabular}{l|cc|c|c|c}
\toprule
\multirow{2}{*}{\textbf{Method}} &
\multicolumn{2}{c|}{\textbf{detection}}&
\multicolumn{1}{c|}{\textbf{map}} &
\multicolumn{1}{c|}{\textbf{track}} &
\multicolumn{1}{c}{\textbf{motion}} \\

& 
\multicolumn{1}{c}{mAP$\uparrow$} &
\multicolumn{1}{c|}{NDS$\uparrow$} & 
\multicolumn{1}{c|}{mAP$\uparrow$} &
\multicolumn{1}{c|}{AMOTA$\uparrow$} & 
\multicolumn{1}{c}{minADE$\downarrow$} \\
\midrule

UniAD~\cite{UniAD}               & 0.380 & 0.359 & -     & -     & -    \\
VAD~\cite{VAD}                   & 0.276 & 0.397 & 0.476 & -     & -    \\
SparseDrive-S~\cite{SparseDrive} & 0.418 & 0.525 & 0.551 & 0.386 & 0.62 \\
DiFSD~\cite{DiFSD}               & 0.410 & 0.528 & 0.560 & -     & -    \\
\midrule
HiP-AD                           & \textbf{0.424} & \textbf{0.535} & \textbf{0.571} & \textbf{0.406} & \textbf{0.61} \\
\bottomrule
\end{tabular}
}
\caption{Comparison of perception, mapping, tracking, and motion prediction performance on the nuScenes validation dataset.}
\label{tab:nuscenes2}
\vspace{-0.1cm}
\end{table}

\noindent
\textbf{Control.}
Similar to CarLLaVA~\cite{CarLLaVA}, we employ spatial waypoints for lateral control. 
For longitudinal control, we first evaluate the speed of the computed driving-style waypoints to ensure their alignment with the predefined velocity ranges associated with the selected driving style. If the speeds are consistent, the driving-style waypoints are used to control the ego vehicle; otherwise, longitudinal control reverts to temporal waypoints.

\subsection{Loss Functions}
HiP-AD can be end-to-end trained and optimized in a fully differentiable manner.
The overall optimization function includes four primary tasks (detection, motion prediction, mapping, and planning). 
Each primary task can be optimized using both classification and regression losses with corresponding weight. 
The overall loss function can be formulated as follows:
\begin{equation}
    \mathcal{L}_{overall} = \mathcal{L}_{det} + \mathcal{L}_{motion} + \mathcal{L}_{map} + \mathcal{L}_{plan}. 
\end{equation}
The planning loss consists of multi-granularity waypoints regression loss and classification loss with modality and driving-style:
\begin{equation}
    \mathcal{L}_{plan} = \sum^{N_g}\mathcal{L}^{j}_{reg}+\mathcal{L}^{m}_{cls}+\mathcal{L}^{d}_{cls},
\end{equation}
$N_g$ is the number of granularity.
We do not use any constraint losses~\cite{VAD, VADv2} nor de-noising tricks~\cite{DiFSD} in planning.

\begin{table}[t]\footnotesize
\centering
\scalebox{0.87}
{
\begin{tabular}{l|cc|cc}
\toprule
\multirow{2}{*}{\textbf{Variants}} &
\multirow{2}{*}{\textbf{PDA}} &
\multirow{2}{*}{\textbf{MG}} &
\multicolumn{2}{c}{\textbf{Closed-loop}} \\

&  &  &
\multicolumn{1}{c}{Driving Score$\uparrow$} &
\multicolumn{1}{c}{Success Rate(\%)$\uparrow$} \\
\midrule
\multirow{1}{*}{Sequential}           
& \checkmark & \checkmark &  73.2 & 45.5   \\

\midrule
\multirow{4}{*}{Unified}  
&    -       &     -      &  41.3 & 16.4 \\
& \checkmark &     -      &  49.7 & 25.5 \\
&    -       & \checkmark &  76.4 & 47.2 \\
& \checkmark & \checkmark &  \textbf{88.3} & \textbf{72.7} \\
\bottomrule
\end{tabular}
}
\caption{Ablation study of the architecture and proposed modules. PDA: Planning Deformable Attention; MG: Multi-Granularity.}
\label{tab:ablation_architecture}
\vspace{-0.5cm}
\end{table}
\begin{table*}[tb]\footnotesize
\centering
\scalebox{0.82}
{
\begin{tabular}{c|cc|cccccc|cc|ccc}

\toprule
\multirow{3}{*}{\textbf{Index}} &
\multirow{3}{*}{\textbf{Fusion}}&
\multirow{3}{*}{\textbf{Align-Matching}}&
\multicolumn{6}{c|}{\textbf{Multi-Granularity Details}} &
\multicolumn{2}{c|}{\textbf{Control}} &
\multicolumn{3}{c}{\textbf{Closed-loop}} \\

& & &
\multicolumn{2}{c|}{Temporal} &
\multicolumn{2}{c|}{Spatial} &
\multicolumn{2}{c|}{Driving Style} &
\multirow{2}{*}{Lon.} &
\multirow{2}{*}{Lat.} &
\multirow{2}{*}{Driving Score$\uparrow$} &
\multirow{2}{*}{Success Rate(\%)$\uparrow$} &
\multirow{2}{*}{Time Out(\%)$\downarrow$} \\

& & &
\multicolumn{1}{c}{2Hz} &
\multicolumn{1}{c|}{5Hz} &
\multicolumn{1}{c}{5m}  &
\multicolumn{1}{c|}{2m}  &
\multicolumn{1}{c}{sty-2Hz}  &
\multicolumn{1}{c|}{sty-5Hz} &
& & & \\

\midrule
1 &           &           &\checkmark&          &          &          &          &          &2Hz     &2Hz&  49.7 &  25.5  & 41.8\\
2 &           &           &\checkmark&          &\checkmark&          &          &          &2Hz     &5m &  53.6 &  29.0  & 27.3\\
3 &\checkmark &           &\checkmark&          &\checkmark&          &          &          &2Hz     &5m &  62.4 &  29.0  & 20.0\\
4 &\checkmark &\checkmark &\checkmark&          &\checkmark&          &          &          &2Hz     &5m &  79.5 &  56.3  & 5.5\\
5 &\checkmark &\checkmark &          &\checkmark&          &\checkmark&          &          &5Hz     &2m &  82.1 &  60.0  & 3.6\\
6 &\checkmark &\checkmark &\checkmark&\checkmark&\checkmark&\checkmark&          &          &5Hz     &2m &  84.2 &  65.5  & 1.8\\
7 &\checkmark &\checkmark &\checkmark&\checkmark&\checkmark&\checkmark&\checkmark&\checkmark&sty-5Hz &2m &  \textbf{88.3} &  \textbf{72.7} & \textbf{1.8} \\
\bottomrule
\end{tabular}
}
\caption{Ablation study of multi-granularity planning. Lon. and Lat. refer to longitudinal and lateral control.}
\label{tab:ablation_planning}
\vspace{-0.2cm}
\end{table*}

\section{Experiments}

\subsection{Dataset and Metrics}
\noindent
\textbf{Dataset.}
In order to comprehensively evaluate the performance of the end-to-end autonomous driving algorithm, our experiments are mainly conducted on challenging closed-loop benchmark Bench2Drive~\cite{bench2drive} dataset, which collects 1000 short clips uniformly distributed under 44 interactive scenarios in CARLA v2~\cite{Carla}. 
There are 950 clips for training and 50 clips for open-loop validation. 
Furthermore, it provides closed-loop evaluation protocols under 220 test routes for fair comparison.
We also evaluate the open-loop performance on the realistic dataset nuScenes~\cite{nuscenes}, which comprises 1,000 videos divided into training, validation, and testing sets with 700, 150, and 150 videos, respectively.

\noindent
\textbf{Metrics.}
For closed-loop evaluation, we employ the official evaluation metrics recommended in Bench2Drive: Driving Score (DS), Success Rate (SR), Efficiency (Eff.), and Comfortness (Com.).
For open-loop evaluation in both dataset, we follow previous works~\cite{UniAD, VAD}, using L2 Displacement Error (L2) and Collision Rate (CR) to measue the planning trajectory.
The results of 3D object detection, tracking, and online mapping on nuScenes dataset are also reported with commonly used metrics~\cite{nuscenes, BEVFormer, MapTR}.

\subsection{Implementation Details}
We adopt ResNet50 as the backbone with 6 decoder layers, utilizing an input resolution of $640\times352$ in Bench2Drive, which serves as the default dataset for our experiments. 
We establish fixed quantities for hybrid task queries, consisting of 900 agents, 100 maps, and 480 planning queries. Each planning query includes 48 modalities, with 10 granularities per modality. 
These granularities include spatial waypoints sampled at uniform intervals of 2m and 5m, temporal waypoints sampled at frequencies of 2Hz and 5Hz, and manual divisions of driving styles across three speed ranges: $[0, 0.4)$, $[0.4, 3)$, and $[3, 10)$ m/s, each with two frequency settings.
Target points and high-level commands are embedded into the planning head through MLPs, while the ego status is excluded from inputs.

The training process consists of two phases. 
Initially, we disable the driving-style head for 12 epochs, followed by 6 epochs of fine-tuning with the driving-style head enabled.
We train the model on 8 NVIDIA 4090 GPUs with a total batch size of 32.
The AdamW optimizer and Cosine Annealing scheduler are utilized with an initial learning rate of $2 \times 10^{-4}$ and a weight decay of 0.01.
Training on the nuScenes dataset using a similar process. Additional implementation details are provided in the appendix.

\subsection{Main Results}

\noindent
\textbf{Bench2Drive.}
We evaluate HiP-AD against state-of-the-art end-to-end autonomous driving methods on the Bench2Drive dataset, with results presented in~\cref{tab:bench2dirve1}.
HiP-AD achieves the best closed-loop performance, demonstrating superior Driving Score and Success Rate. Compared to the second-place method~\cite{DriveTransformer}, it significantly improves by over 20\% in Driving Score and 30\% in Success Rate. Furthermore, HiP-AD attains a comparable L2 error score when compared to other leading methods.
The primary limitation of our approach is in Comfortness. However, we emphasize that comparing Comfortness is meaningful only among methods with similar Success Rate scores. Behaviors such as sudden braking or turning, while potentially reducing Comfortness, are often necessary to ensure the successful completion of the evaluation.

Additionally, we present the multi-ability scores in~\cref{tab:bench2dirve2}, highlighting HiP-AD's exceptional performance across diverse driving scenarios. HiP-AD significantly enhances capabilities in scenarios such as Merging, Overtaking, Emergency Brake, and Taffic Sign, leading to an overall score improvement of over 25\%.
Give Way illustrates some rare driving scenarios that require ego vehicle yield to an emergency vehicle coming from behind.
Considering the imbalance datasets, making quick decisions is still a challenge for E2E-AD methods.

\noindent
\textbf{NuScenes.}
We further evaluate HiP-AD's performance on the open-loop dataset nuScenes, focusing on perception and motion prediction tasks. As shown in~\cref{tab:nuscenes1}, HiP-AD achieves the lowest Collision Rate among all compared methods while maintaining a competitive L2 error.
The perception and prediction results, presented in~\cref{tab:nuscenes2}, demonstrate that HiP-AD delivers strong performance in both tasks, highlighting the robustness and effectiveness of the proposed unified framework.

\subsection{Ablation Study}
In the ablation experiments, we use a small test set of Bench2Drive to save computational resources.
This subset consists of 55 routes (25\% of the total 220 routes), with 44 routes selected one-to-one from 44 unique scenarios and the remaining 11 routes randomly chosen from rest routes.

\noindent
\textbf{Effect of architecture and modules.}
As shown in ~\cref{tab:ablation_architecture}, we evaluate the contributions of planning deformable attention and multi-granularity representation. The results indicate that both components play a critical role in enhancing overall performance, with multi-granularity particularly noteworthy for its provision of improved control.
Additionally, we compare the proposed unified framework with its sequential variant. In the sequential version, the perception components of HiP-AD are executed first, followed by the planning components. In contrast, the unified version runs perception and planning iteratively in parallel. Experimental results reveal that the unified version significantly outperforms the sequential variant, demonstrating the superiority of the unified framework.

\noindent
\textbf{Effect of multi-granularity planning.}
\cref{tab:ablation_planning} presents the ablation study on multi-granularity planning query design. The 1st setting employs only 2Hz temporal waypoints, consistent with VAD \cite{VAD} or UniAD \cite{UniAD}. The 2nd setting incorporates both temporal and spatial waypoints, similar to CarLLaVA \cite{CarLLaVA}. The 3rd  and 4th setting use the same waypoints as the 2nd but introduces granularity fusion and align-matching, demonstrating significant performance improvements.
The 5th setting utilizes 5Hz and 2m waypoints, highlighting that higher-frequency waypoints enhance fine-grained control. The 6th setting combines dense and sparse sampling intervals, showing that granularity fusion of sampling intervals boosts performance. The 7th setting integrates driving style, achieving a 7\% improvement in Success Rate and delivering the best overall performance.
The waypoints used for control correspond to the granularity settings, as detailed in \cref{tab:ablation_planning}. 
Additionally, the hesitation phenomenon is measured by ``Agent timed out" status, which shows the multi-granularity planning not only maintain safe driving, but also encourages behavior learning.

\begin{figure}[t]
  \centering
  \vspace{0.1cm}
  \includegraphics[width=1\linewidth]{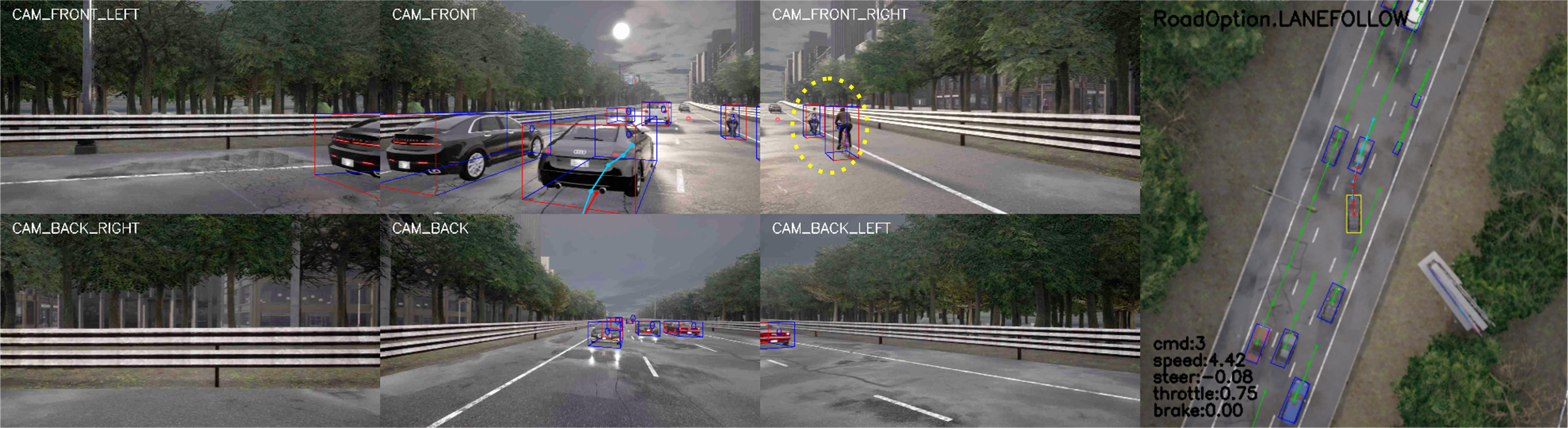}
  \includegraphics[width=1\linewidth]{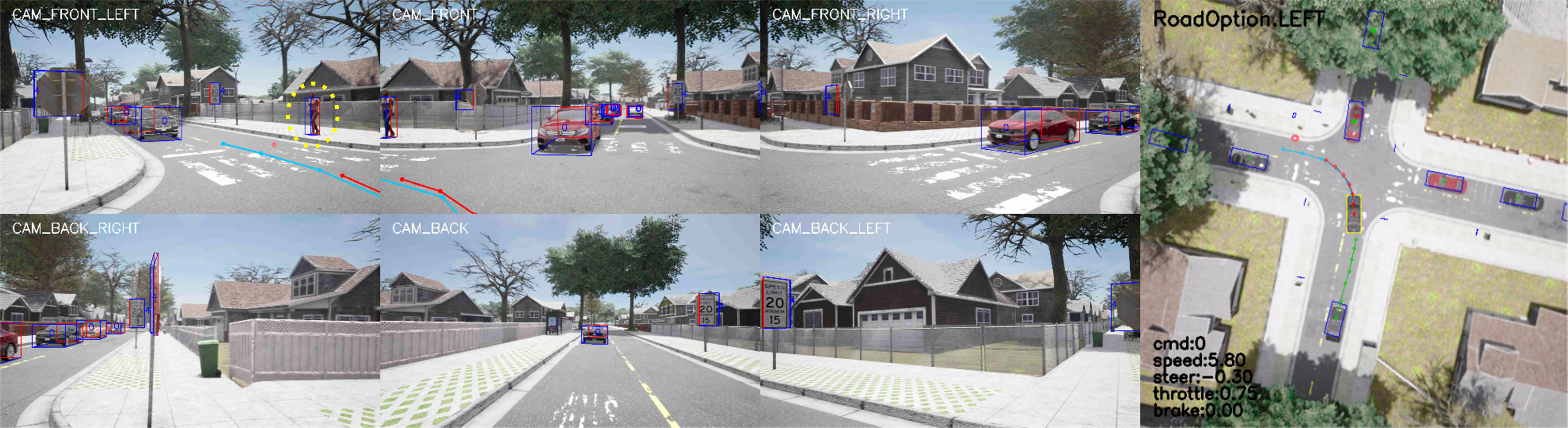}
  \caption{Qualitative results of HiP-AD on closed-loop routes, including perception, motion, and planning trajectories. Spatial waypoints are colored in {\color{skyblue}skyblue}, while driving-style waypoints are colored in \hlr{red}. Important objects are highlight in \hly{yellow} circle.
  }
  \label{fig:7_qualitative_results}
  \vspace{-0.5cm}
\end{figure}

\subsection{Qualitative Results}
We visualize the closed-loop results on the Bench2Drive test routes, as shown in ~\cref{fig:7_qualitative_results}. Two typical scenarios are illustrated: the first involves an unprotected left turn, and the second features a hazard in the roadside lane.
In the first scenario, guided by spatial waypoints, the ego vehicle successfully executes a left turn, adhering to the target trajectory and changing lanes to avoid the bicycle. In the second scenario, the ego vehicle decelerates in accordance with the proposed driving-style waypoints upon detecting obstacles, such as a pedestrian on the side and a car ahead.
These results demonstrate the effectiveness of HiP-AD in handling complex driving scenarios. Additional qualitative results are provided in the supplementary materials.

\vspace{-0.1cm}
\section{Conclusion and Limitation}
In this paper, we introduce HiP-AD, a novel end-to-end autonomous driving framework. HiP-AD features a unified decoder capable of simultaneously executing perception, prediction, and planning tasks. The planning query iteratively interacts with perception features in BEV space and multi-view image features in the perspective view, enabling comprehensive interaction. Additionally, we propose a multi-granularity planning strategy that integrates diverse planning trajectories with rich supervision to enhance ego-vehicle control.
Extensive experiments on both closed-loop and open-loop datasets demonstrate outstanding planning performance, verifying the effectiveness of HiP-AD in complex driving scenarios.

Limitations: Despite achieving excellent performance in both closed-loop and open-loop evaluations, extensive real-world testing is still necessary. 
Furthermore, avoiding collisions with vehicles rapidly approaching from behind remains a challenge. 
These issues will be a focus of our future research.

{
    \small
    \bibliographystyle{ieeenat_fullname}
    \bibliography{main}
}

\newpage
\clearpage
\maketitlesupplementary
\setcounter{page}{1}
\setcounter{section}{0}
\setcounter{figure}{0}
\setcounter{table}{0}
\setcounter{equation}{0}

\appendix

\section{Comparison with Previous E2E-AD Methods}
We summarize two key differences between HiP-AD and previous E2E-AD methods, primarily in planning query design and planning interaction, as illustrated in ~\cref{tab:comparison}.
Compared to previous E2E-AD models, we introduce driving-style waypoints prediction to enhance the precision of autonomous vehicle steering and acceleration/deceleration. Furthermore, we employ a multi-granularity strategy to provide complementary long-short information, enriching the addtional supervision.
Moreover, we develop a comprehensive interaction mechanism in which the planning query not only interacts with perception tasks but also engages with sparse image features through planning deformable attention, facilitating the learning of unannotated scene information.
\begin{table*}[t]\footnotesize
\centering
\scalebox{0.92}
{
\begin{tabular}{l|cccc|cccc}
\toprule
\multirow{3}{*}{\textbf{Method}} &
\multicolumn{4}{c|}{\textbf{Planning waypoints}} & \multicolumn{4}{c}{\textbf{Planning interaction}} \\ 
\cmidrule{2-9} 
& \multicolumn{3}{c|}{heterogeneours waypoints} & \multirow{2}{*}{multi-granularity} & 
\multirow{2}{*}{perception} & \multicolumn{3}{|c}{scenes representations}

\\

 & default / temporal & spatial & \multicolumn{1}{c|}{driving-style} &  &  & \multicolumn{1}{|c}{BEV} & global images & sparse images
\\
\midrule
UniAD~\cite{UniAD}               & \checkmark & & & & \checkmark & & & \\
VAD~\cite{VAD}                   & \checkmark & & & & \checkmark & & & \\
Para-Drive~\cite{PARA-Drive}     & \checkmark & & & & & \checkmark & & \\ 
GenAD~\cite{GenAD}               & \checkmark & & & & & \checkmark & & \\
SparseDrive~\cite{SparseDrive}   & \checkmark & & & & \checkmark & & & \\ 
DiFSD~\cite{DiFSD}               & \checkmark & & & & \checkmark & & & \\
DriveTransformer~\cite{DriveTransformer} 
                                 & \checkmark & & & & \checkmark &  & \checkmark & \\
CarLLaVA~\cite{CarLLaVA}         & \checkmark & \checkmark & & & \checkmark & & \checkmark & \\                       
\midrule                                
HiP-AD (ours)                    & \checkmark & \checkmark & \checkmark & \checkmark & \checkmark & & & \checkmark \\
\bottomrule
\end{tabular}
}
\caption{Comparison of representative related models and our HiP-AD in terms of planning query design and planning interaction.}
\label{tab:comparison}
\end{table*}

\section{More Implementation Details}

\noindent
\textbf{Supervision.}
All ground truth (GT) of multi-granularity waypoints are derived from the future trajectory of the ego vehicle, differing primarily in the sampling method or strategy.
For example, we gather all future locations of the ego car within a single video and apply linear fitting to obtain a trajectory function.
Then, we can sample from this function at any equal distance intervals to attain the ground truth of spatial waypoints.
In contrast, temporal and driving-style predictions share the same ground truth of temporal waypoints, which is sampled directly from future locations at equal time intervals.

In addition to directly supervising perception and trajectory, we also incorporate additional supervisory signals to enhance the training process. The auxiliary supervision includes a sparse depth map and ego status, which can be formulated as follows:
\begin{equation}
\mathcal{L}_{aux} = \mathcal{L}_{depth} + \mathcal{L}_{status}.
\end{equation}

\noindent
\textbf{Ablation Routes.}
There are 220 routes in the Bench2Drive closed-loop test routes, officially divided into 44 scenarios based on various weather conditions and scene events.
It is computationally exhaustive for us to conduct all experiments on the total number of test routes.
Therefore, to conserve computational resources, we utilize a small test set on Bench2Drive for our closed-loop ablation experiments.
It consists of 55 routes (25\% of 220) which can be divided into two parts.
The first part includes 44 routes, selected one-to-one from each of the 44 scenarios, while the remaining 11 routes are chosen randomly.
This small test set not only includes all scenarios but also maintains the distribution of the total test routes for comprehensive evaluation.
We list route IDs in~\cref{tab:supp_route_ids}.
\begin{table*}[h]\footnotesize
\centering

\begin{tabular}{l|c}
\toprule
Scenarios & Route IDs\\
\midrule
\multirow{4}{*}{44 scenarios}  
&1711, 1773, 1790, 1825, 1852, 1956, 2050, 2082, 2084, 2115, 2127,\\
&2144, 2164, 2201, 2204, 2273, 2286, 2373, 2390, 2416, 2509, 2534, \\
&2664, 2709, 2790, 3086, 3248, 3364, 3436, 3464, 3540, 3561, 3936, \\
&14194, 14842, 17563, 17752, 23658, 23695, 23771, 23901, 26458, 28087, 28099\\
\midrule
\multirow{1}{*}{11 random routes}  
&1792, 2086, 2129, 2283, 2539, 2668, 26406, 26956, 27494, 27532, 28154\\
\bottomrule
\end{tabular}
\caption{A small test set of Bench2Drive for ablation experiments.}
\label{tab:supp_route_ids}
\end{table*}

\noindent
\textbf{Speed Setting.}
The speed of driving-styles is empirically divided into three intervals: [0, 0.4), [0.4, 3), and [3, 10). 
Each interval corresponds to different driving scenarios.
\begin{itemize}
    \item The interval 0-0.4 corresponds to parking maneuvers.
    \item The interval 0.4-3 corresponds to low-speed actions, such as strategic lane changes and stop-and-go situations.
    \item The interval 3-10 corresponds to normal driving. 
\end{itemize}

\noindent
\textbf{NuScenes Training.}
The training parameters on nuScenes dataset are similar to those used in Bench2Drive, except that the resolution is changed to $704\times256$.
Moreover, the training process is different from Bench2Drive, considering the different intentions between open-loop and closed-loop evaluations.
First, since nuScenes does not require precise control, we disenable driving-style waypoints during training and use temporal waypoints for performance evaluation.
Second, as a multi-task learning framework, we adopt the training approach used in SparseDrive~\cite{SparseDrive} to maximize perception performance, which entails first training the perception component, followed by motion prediction and planning.

\noindent
\textbf{Closed-loop Inference.}
Although we employ waypoints at varying frequencies for supervision, our model is trained and inferenced at 2Hz. 
Considering that the operational frequency during the Bench2Drive closed-loop simulation is 10Hz, we set up multiple memory modules (e.g., 5) to ensure the inference frequency aligns with the training phase at 2Hz.
It means only one memory is available in each timestamp to output temporal task queries and keep top $k$ updated task queries.
Following~\cite{SparseDrive}, we set $k$ as 600 for agent queries, 0 for map queries, and 480 for planning queries.
Additionally, we list the inference time and parameters of the model in~\cref{tab:supp_fps_parameter}.
\begin{table}[h]\footnotesize
\centering
\resizebox{0.9\linewidth}{!}
{
\begin{tabular}{l|c|c|c|c}
\toprule
Dataset & Parameters & GFLOPs & Latency & FPS \\
\midrule
\multirow{1}{*}{nuScenes}  
& 90.0M & 202.9 & 109.9ms & 9.1 \\
\multirow{1}{*}{Bench2Drive}  
& 97.4M & 256.9 & 138.9ms & 7.2 \\
\bottomrule
\end{tabular}
}
\caption{The model is measured on a single NVIDIA 3090 GPU.}
\label{tab:supp_fps_parameter}
\end{table}

\section{More Experiments}

\noindent
\textbf{Ganularity Selection.}
\begin{table}[tb]\footnotesize
\centering
\scalebox{1}
{
\begin{tabular}{cc|cc}

\toprule
\multicolumn{2}{c|}{\textbf{Control}} & \multicolumn{2}{c}{\textbf{Closed-loop Metric}} \\

Lon. &
Lat. &
Driving Score$\uparrow$ &
Success Rate(\%)$\uparrow$ \\

\midrule
2Hz & 5m & 76.35 & 52.72  \\
2Hz & 2m & 79.03 & 56.36  \\
5Hz & 5m & 81.85 & 65.45  \\
5Hz & 2m & \textbf{86.05} & \textbf{69.09}  \\

\bottomrule
\end{tabular}
}
\caption{Ablation study on controlling ego-vehicle under different configurations.}
\label{tab:supp_action_control}
\end{table}
We output spatial waypoints at dense and sparse distance intervals, along with high-frequency and low-frequency temporal waypoints, as well as driving-style waypoints.
Therefore, we conduct an ablation study to explore which configuration yields better vehicle control in closed-loop system.
As shown in the~\ref{tab:supp_action_control}, there are four combinations: dense and sparse intervals, as well as high and low frequencies.
Among these, the sparse-interval and low-frequency combination perform the worst, while the dense-interval and high-frequency combination provide the best vehicle control. 
Therefore, we chose the dense-interval and high-frequency combination as the final control method for our model.
It is worth noting that even the worst combination our approach still outperforms the previous state-of-the-art methods. 
This clearly demonstrates the superiority of our method.

\noindent
\textbf{Number of Modalities.}
\begin{table}[tb]\footnotesize
\centering
\resizebox{0.8\linewidth}{!}
{
\begin{tabular}{c|cccc}
\toprule
\multirow{2}{*}{\textbf{Number}} & \multicolumn{4}{c}{\textbf{Closed-loop Metric}}  \\
& DS $\uparrow$  & SR(\%) $\uparrow$ & Eff. $\uparrow$ & Com. $\uparrow$\\ \midrule                           
18 & 86.56 & 62.27 & 197.21 & 17.35   \\
48 & \textbf{86.77} & \textbf{69.09} & \textbf{203.12} & \textbf{19.36}   \\
\bottomrule
\end{tabular}
}
\caption{Comparison of the number of modalities: `Eff.' and `Com.' are abbreviations for Efficiency and Comfort.}
\label{tab:supp_num_modality}
\end{table}
Previous methods~\cite{SparseDrive} usually use 18 modalities in Bench2Drive. We provide our 18 modalities results in ~\cref{tab:supp_num_modality}. The Driving Score, Efficiency and Comfortness are similar on modalities 18 and 48. The 48 modalities version improves over 6\% on Success Rate, which indicates increasing modalities number also contribute to final performance. The 18 modalities version is still much better than other methods, which indicates  modality number is not the key impact for performance.

\noindent
\textbf{Sparse Interaction.}
We compare the performance of planning queries with global and sparse interactions of image features.
In the Global setting, considering memory constraints, we employ global attention, where the planning query interacts with 1/16 downsample image features.
In contrast to Sparse setting, we utilize the proposed planning deformable attention to dynamically sample image features around trajectory points for interaction.
As shown in~\cref{tab:supp_sparse_global}, the Sparse setting achieve better driving, demonstrating that sparse local interactions can effectively learn latent scene representations, which is more beneficial for closed-loop systems.
\begin{table}[tb]\footnotesize
\centering
\resizebox{1\linewidth}{!}
{
\begin{tabular}{l|c|cc}

\toprule
\multirow{2}{*}{\textbf{Setting}}&\multirow{2}{*}{\textbf{Interaction}}&\multicolumn{2}{c}{\textbf{Closed-loop Metric}}\\
 &  & 
Driving Score$\uparrow$ &
Success Rate(\%)$\uparrow$ \\

\midrule
Global & Cross Attention & 73.4 & 49.1  \\
Sparse & Deformable Attention & \textbf{84.2} & \textbf{65.5} \\
\bottomrule
\end{tabular}
}
\caption{Comparison of Global and Sparse Interactions.}
\label{tab:supp_sparse_global}
\vspace{-0.5cm}
\end{table}

\section{More Qualitative Analysis}
We present additional visualization results of our HiP-AD to demonstrate its effectiveness in both open-loop and closed-loop evaluations.

\subsection{Open-Loop}
\begin{figure*}[htbp]
    \centering
    \begin{subfigure}[b]{0.88\textwidth}
        \centering
        \includegraphics[width=\textwidth]{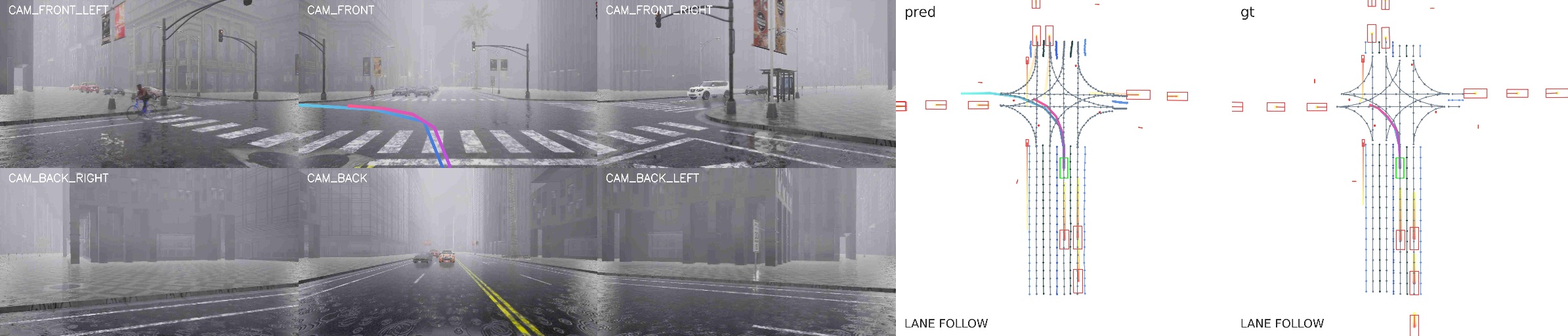}
        \includegraphics[width=\textwidth]{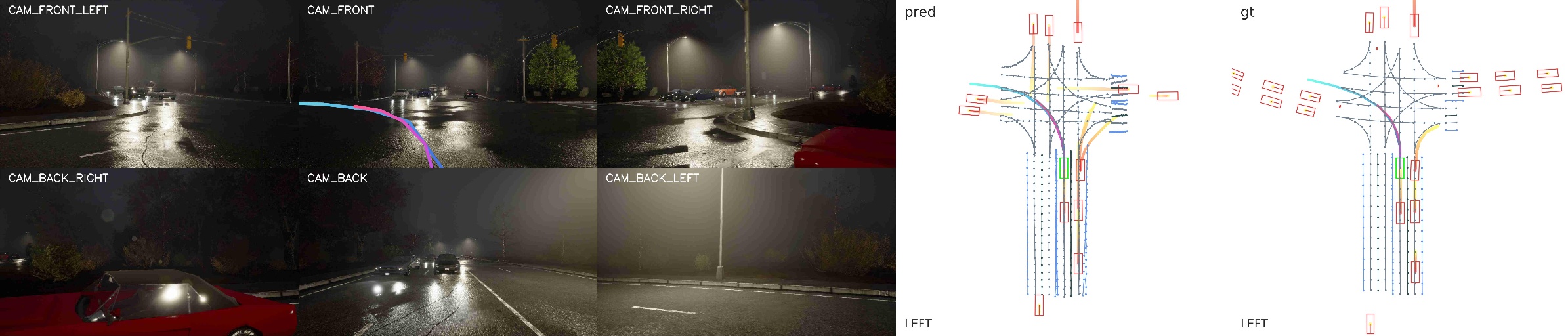}
        \caption{Left turn at an intersection.}
        \label{fig:supp_open_b2d_a}
    \end{subfigure}
    \vfill
    \begin{subfigure}[b]{0.88\textwidth}
        \centering
        \includegraphics[width=\textwidth]{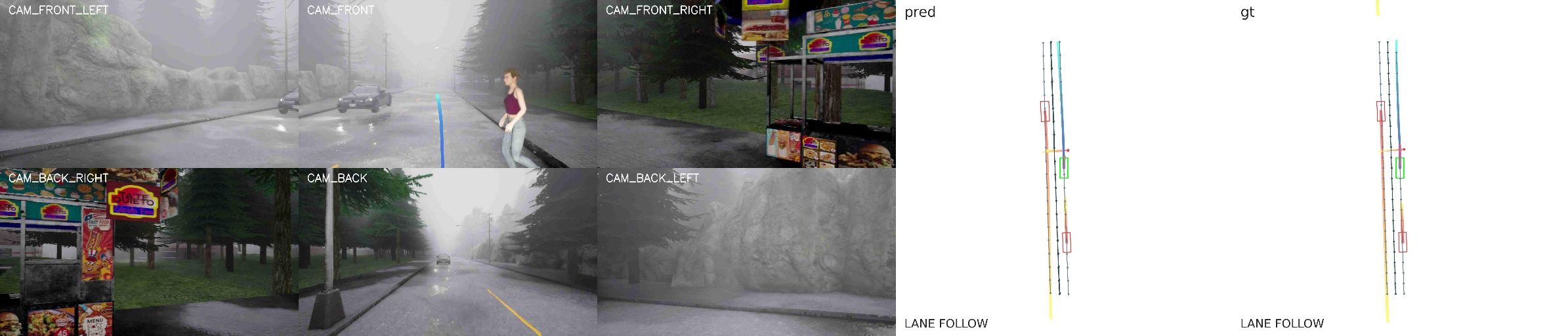}
        \includegraphics[width=\textwidth]{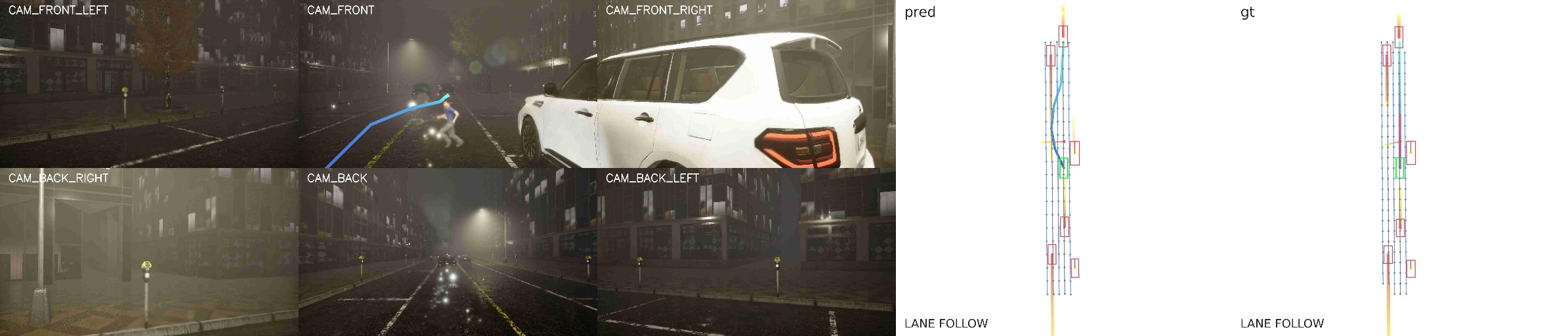}
        \caption{Give way to pedestrians.}
        \label{fig:supp_open_b2d_b}
    \end{subfigure}
    \vfill
    \begin{subfigure}[b]{0.88\textwidth}
        \centering
        \includegraphics[width=\textwidth]{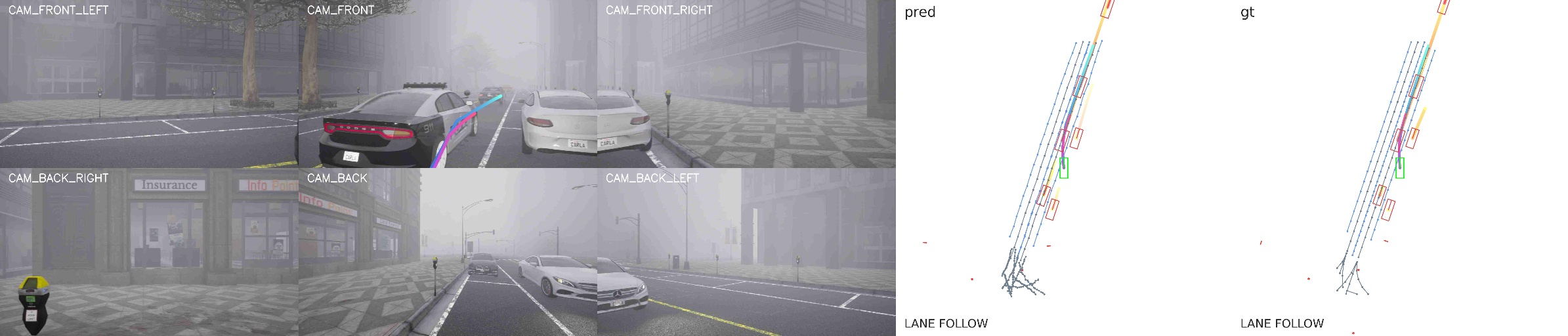}
        \includegraphics[width=\textwidth]{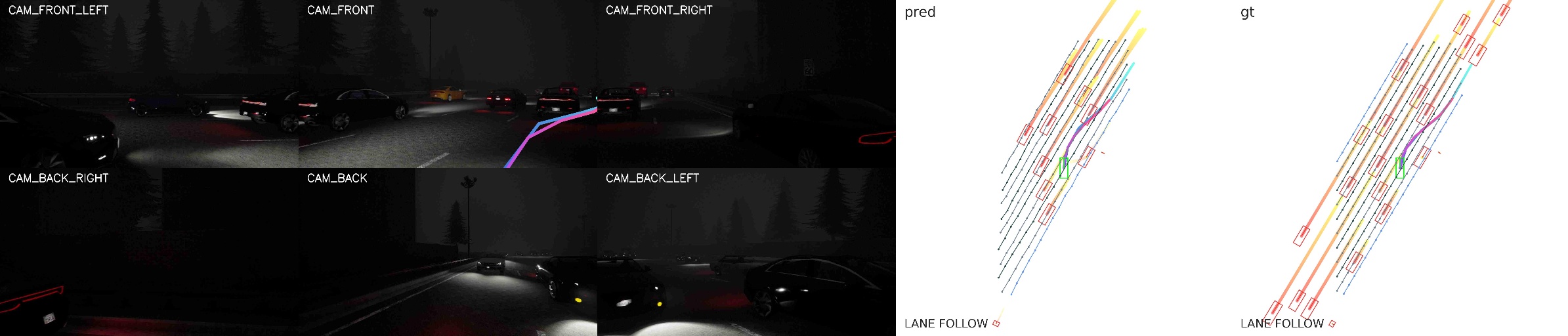}
        \caption{Detour around obstacles.}
        \label{fig:supp_open_b2d_c}
    \end{subfigure}
    \caption{Illustration of open-loop results on Bench2Drive validation dataset.}
    \label{fig:supp_open_b2d}
\end{figure*}

\noindent
\textbf{Bench2Drive.}
As shown in~\cref{fig:supp_open_b2d}, we visualize three driving scenarios on the open-loop validation set of Bench2Drive in both daytime and nighttime.
The three driving scenarios are giving way to pedestrians, left turn at an intersection, and detouring around obstacles.
We use cyan-blue lines represent spatial waypoints, while the purple-red lines indicate driving-style waypoints.

In the first driving scenario~\cref{fig:supp_open_b2d_a}, HiP-AD detects most vehicles at the intersection, reconstructing the map and providing suitable trajectories to ensure that the ego vehicle can successfully make a left turn in the absence of lane markings.
In the second driving scenario~\cref{fig:supp_open_b2d_b}, HiP-AD detects a pedestrian crossing the road in front of the ego vehicle and predicts waypoints for a cautious driving style, reducing speed to avoid a collision with the pedestrian.
The third scenario~\cref{fig:supp_open_b2d_c} showcases HiP-AD's ability to navigate around obstacles, even in complex situations such as starting from a stop or driving at night.

\begin{figure*}[h!]
  \centering
  \includegraphics[width=0.88\textwidth]{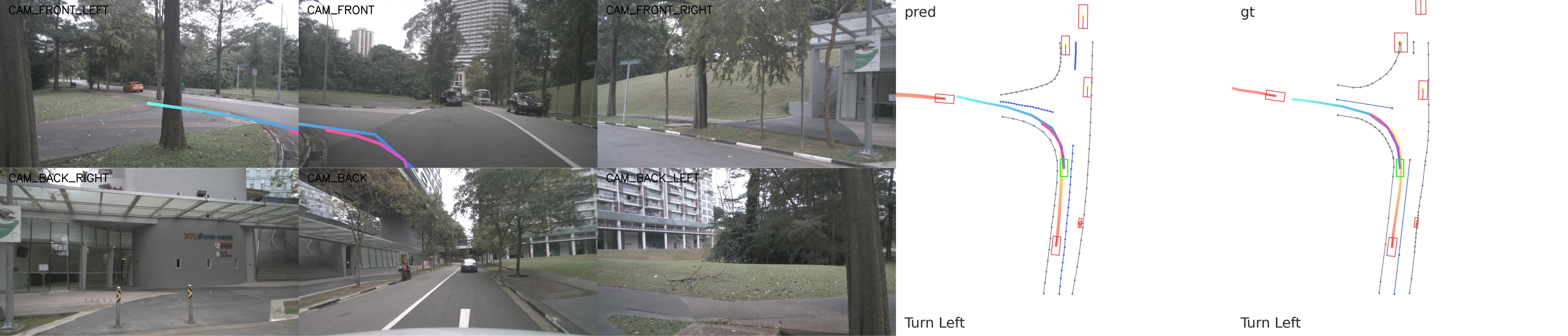}
  \includegraphics[width=0.88\textwidth]{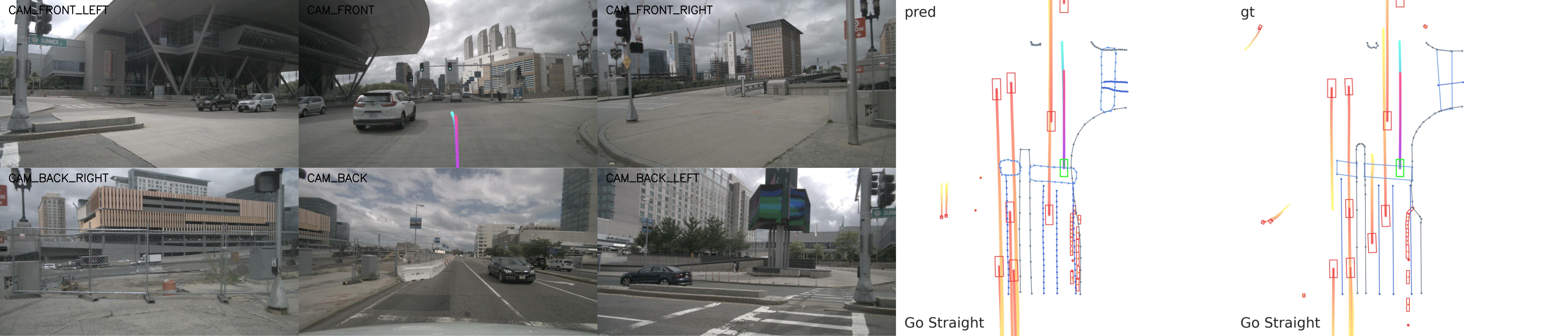}
  \includegraphics[width=0.88\textwidth]{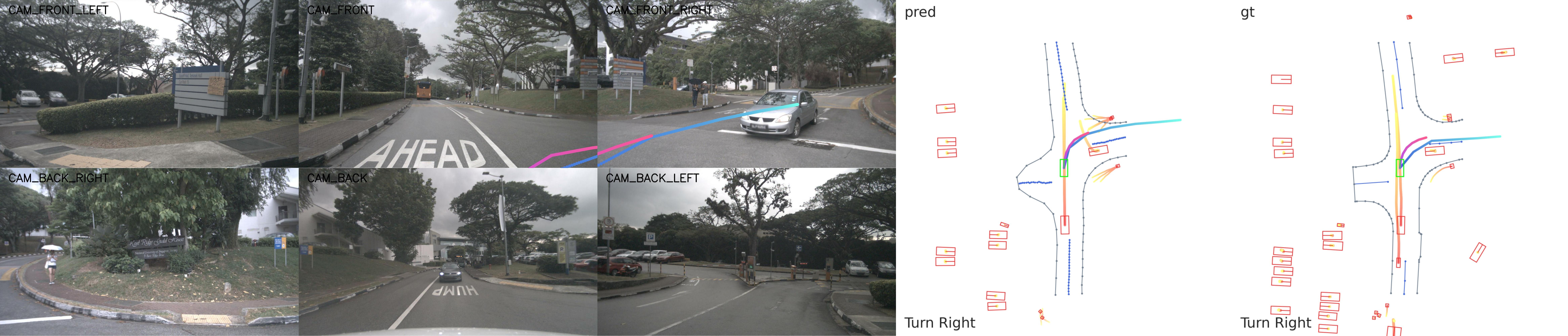}
  \includegraphics[width=0.88\textwidth]{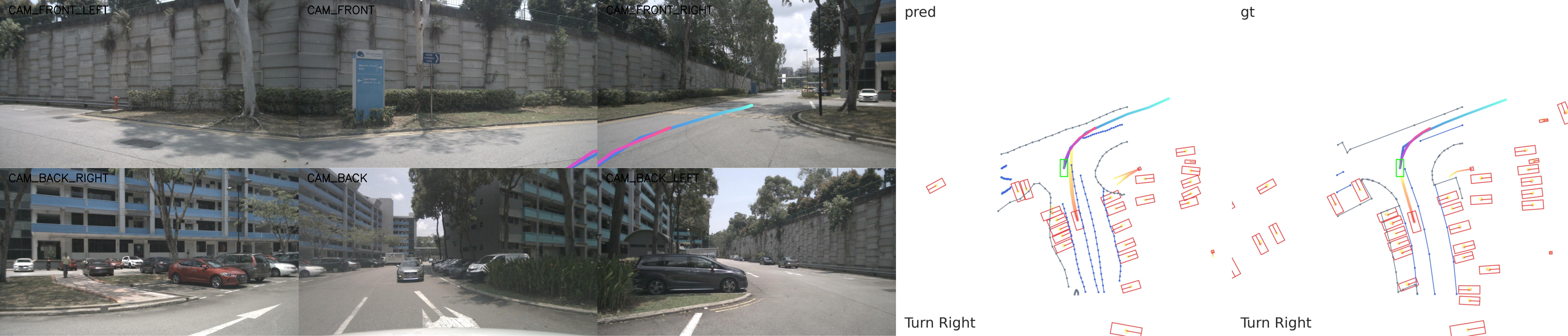}
  \includegraphics[width=0.88\textwidth]{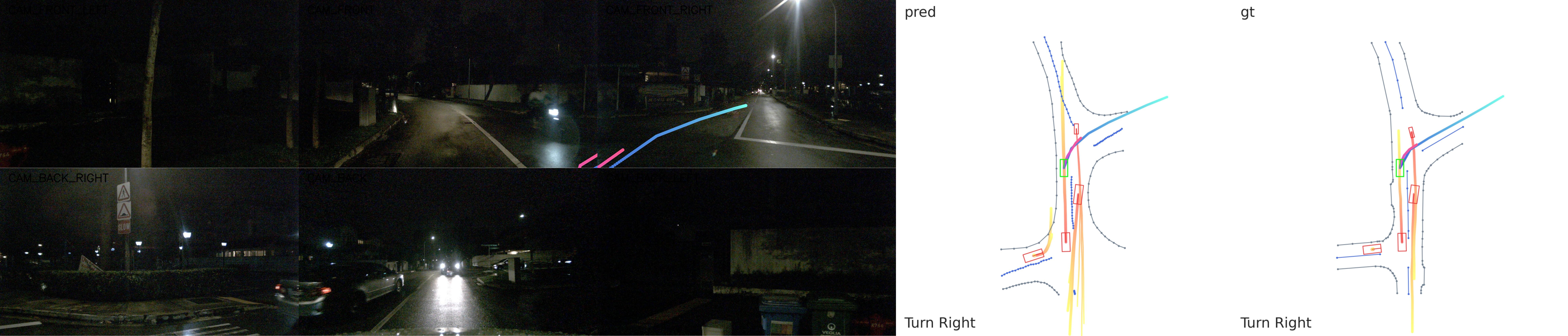}
  \caption{Illustration of open-loop results on nuScenes validation dataset.}
  \label{fig:supp_open_nus}
\end{figure*}

\noindent
\textbf{NuScenes.}
We also provide visualization results on the nuScenes dataset, as shown in~\cref{fig:supp_open_nus}. 
These driving scenarios include left turns, straight driving, and right turns under realistic conditions, such as at intersections or low illumination.
Compared to ground truth, our model effectively locates obstacles, reconstructs map elements, and produces planning results that align closely with ground truth. This demonstrates the robustness of our method in perception and planning within real-world scenarios.

\subsection{Closed-loop}
\begin{figure*}[h!]
  \centering
    \includegraphics[width=0.84\textwidth]{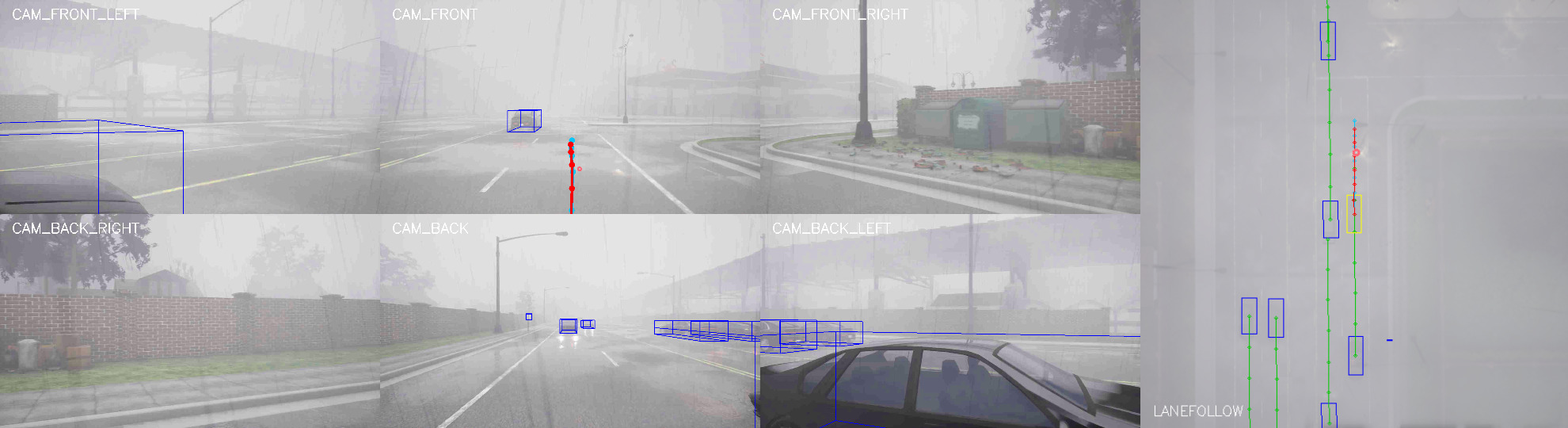}
    \includegraphics[width=0.84\textwidth]{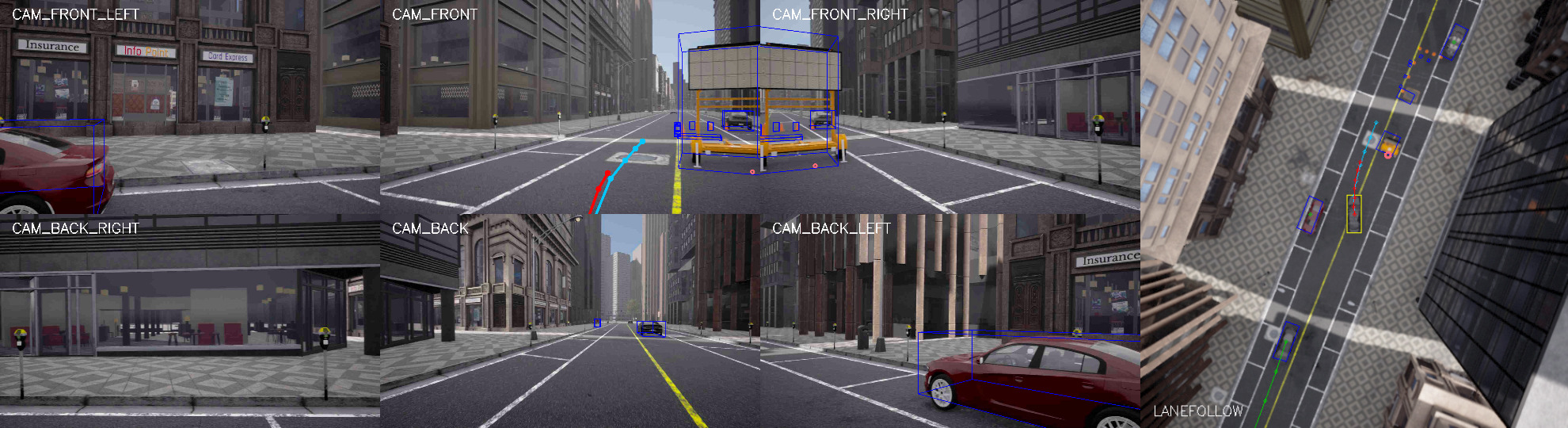}
    \includegraphics[width=0.84\textwidth]{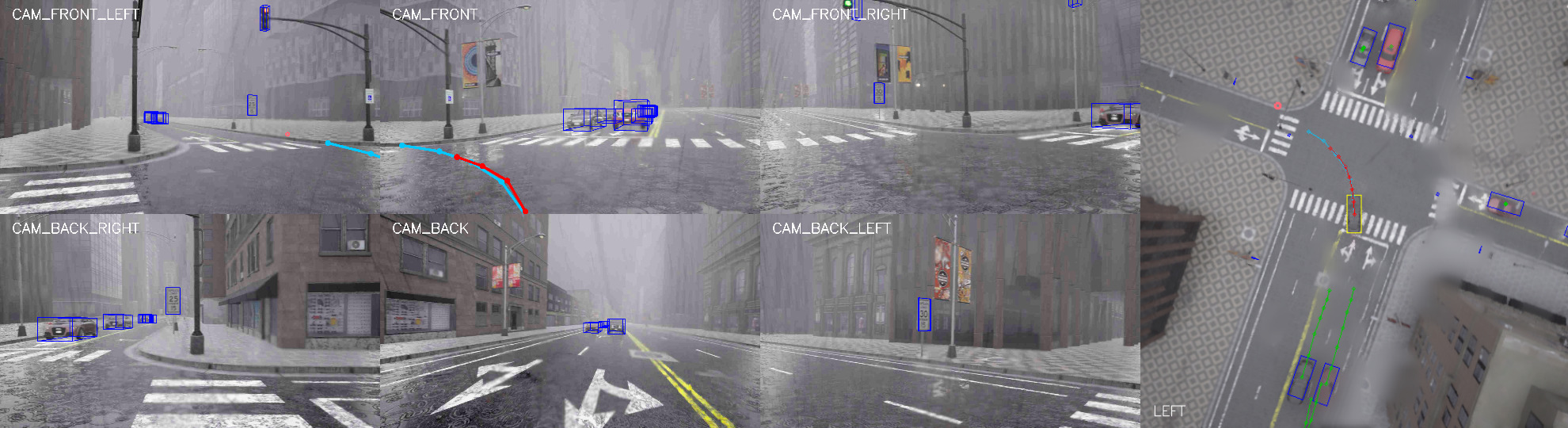}
    \includegraphics[width=0.84\textwidth]{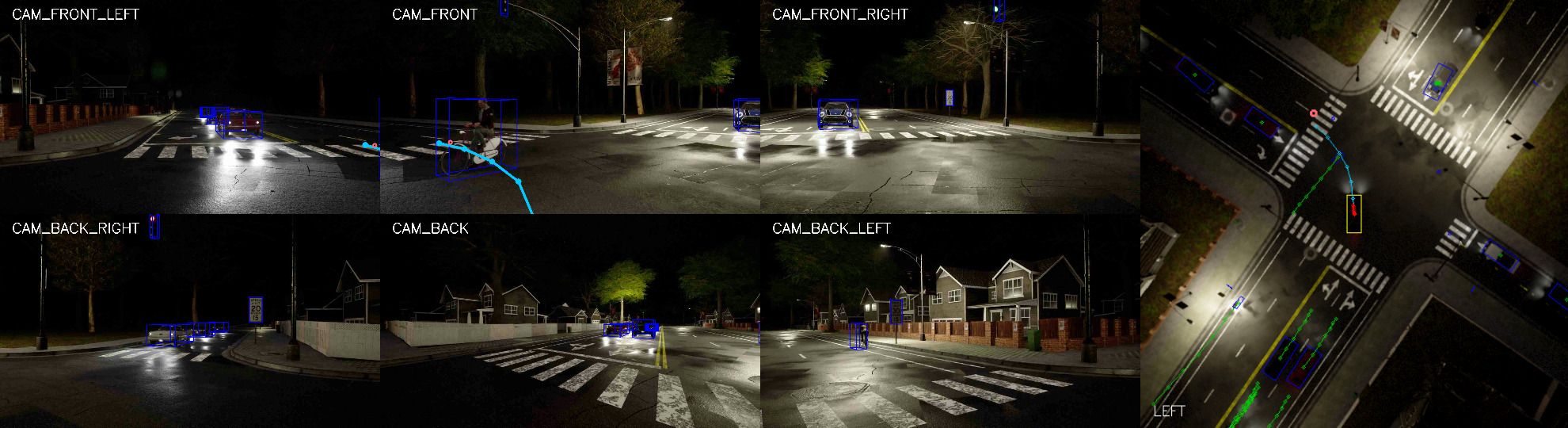}
    \includegraphics[width=0.84\textwidth]{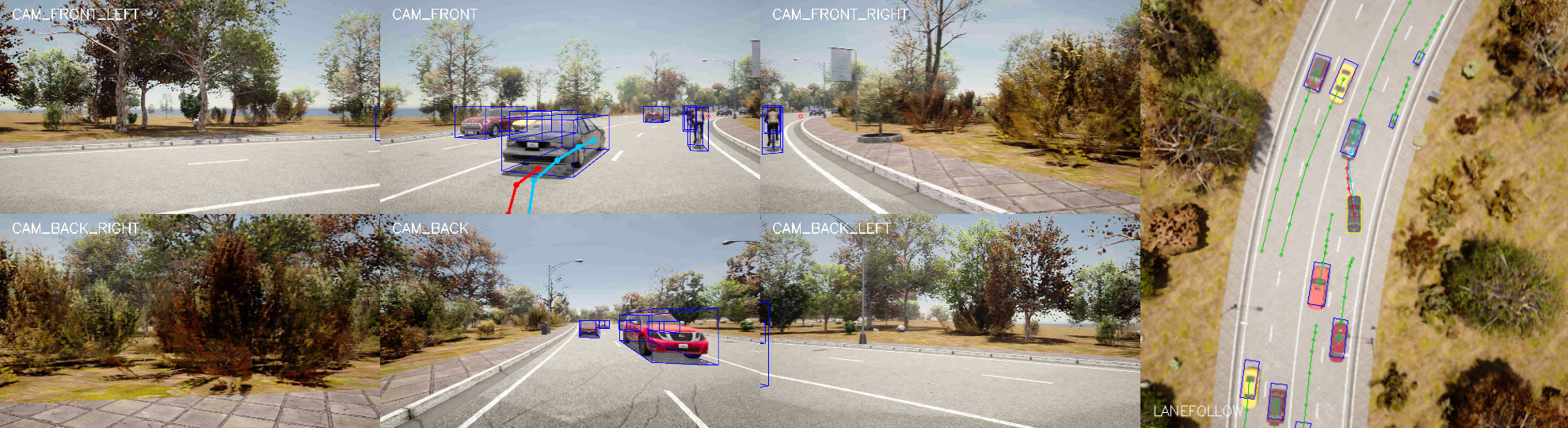}
  \caption{Illustration of closed-loop results on Bench2Drive test routes.}
  \label{fig:supp_close_b2d}
\end{figure*}

\noindent
\textbf{Qualitative Results.}
The closed-loop visualization results of Bench2Drive are shown in ~\cref{fig:supp_close_b2d}. It illustrates the decision-making process and precise control of HiP-AD across various scenarios, including urban streets, intersections, and highways, under conditions such as nighttime and fog.
Thanks to the proposed strategies and the unified architecture, HiP-AD demonstrates strong adaptability and robustness in these closed-loop scenarios.

\end{document}


\maketitle
\input{sec/X_suppl}

{
    \small
    \bibliographystyle{ieeenat_fullname}
    \bibliography{main}
}
